\documentclass[journal]{IEEEtran}

\usepackage{cite}
\usepackage{amsmath,amssymb,amsfonts}
\usepackage{bbm}
\usepackage{algorithmic}
\usepackage{graphicx}
\usepackage{textcomp}

\usepackage{bm}

\usepackage{array}
\usepackage{booktabs}
\usepackage{url}
\usepackage{xcolor}

\usepackage{algorithm}

\usepackage{listings}

\usepackage{cuted}
\definecolor{typeblue}{rgb}{0.10,0.10,0.55} 
\definecolor{lstnum}{rgb}{0.05,0.40,0.50}   
\definecolor{oaext}{rgb}{0.78,0.20,0.04}    
\lstdefinestyle{oajson}{
  basicstyle=\ttfamily\footnotesize,
  keywordstyle=\color{typeblue}\bfseries,
  morekeywords={object,array,integer,string,boolean,number,true,false,null,
                schema,gold,candidate},
  keywords=[2]{idScope,ref,order,align,fixed,score,exact,jaro,invdiff,threshold,
               valueWeight,keyImportance,valueImportance,enum,prefixItems,
               prefixWeights,ignoreExcess},
  keywordstyle=[2]{\color{oaext}\bfseries},
  showstringspaces=false,
  breaklines=true,
  columns=fullflexible,
  keepspaces=true,
  literate=%
    {0}{{\color{lstnum}0}}1 {1}{{\color{lstnum}1}}1 {2}{{\color{lstnum}2}}1
    {3}{{\color{lstnum}3}}1 {4}{{\color{lstnum}4}}1 {5}{{\color{lstnum}5}}1
    {6}{{\color{lstnum}6}}1 {7}{{\color{lstnum}7}}1 {8}{{\color{lstnum}8}}1
    {9}{{\color{lstnum}9}}1,
}
\lstdefinestyle{oajsonschema}{
  basicstyle=\ttfamily\footnotesize,
  keywordstyle=\color{typeblue}\bfseries,
  morekeywords={type,properties,items,schema,gold,candidate,
                entities,relations,id,name,mentions,text,subject,predicate,object,
                events,marker,
                people,companies,employment,acquaintance,partnership,
                mentorships,mentor,mentee,agenda,period,
                root,nodes,concept,attributes,value,
                itinerary,city,days,
                plan,action,args,
                indices,
                title,industry,person,company,role,source,target,relation},
  deletekeywords={array,integer,string,boolean,number,true,false,null},
  keywords=[2]{idScope,ref,order,score,valueWeight,keyImportance,valueImportance,
               enum,prefixItems,prefixWeights,ignoreExcess,threshold},
  keywordstyle=[2]{\color{oaext}\bfseries},
  moredelim=[s][\color{lstnum}]{:"}{"},
  showstringspaces=false,
  breaklines=true,
  columns=fullflexible,
  keepspaces=true,
  literate=%
    {0}{{\color{lstnum}0}}1 {1}{{\color{lstnum}1}}1 {2}{{\color{lstnum}2}}1
    {3}{{\color{lstnum}3}}1 {4}{{\color{lstnum}4}}1 {5}{{\color{lstnum}5}}1
    {6}{{\color{lstnum}6}}1 {7}{{\color{lstnum}7}}1 {8}{{\color{lstnum}8}}1
    {9}{{\color{lstnum}9}}1,
}

\lstdefinestyle{oajsondata}{
  basicstyle=\ttfamily\scriptsize,
  morekeywords={gold,candidate},
  keywords=[2]{people,companies,employment,acquaintance,partnership,
               id,name,title,industry,person,company,role,
               source,target,relation,
               mentorships,mentor,mentee,agenda,period,
               events,marker,
               root,nodes,concept,attributes,value,
               itinerary,city,days,
               plan,action,args,
               indices,
               entities,relations,type,mentions,text,subject,predicate,object},
  keywordstyle=\color{typeblue}\bfseries,
  keywordstyle=[2]\color{typeblue}\bfseries,
  moredelim=[s][\color{lstnum}]{:"}{"},
  showstringspaces=false,
  breaklines=true,
  columns=fullflexible,
  keepspaces=true,
}

\lstdefinestyle{oajsonchanges}{
  basicstyle=\ttfamily\scriptsize,
  keywords={}, keywords=[2]{},
  morekeywords={gold,candidate},
  keywordstyle=\color{typeblue}\bfseries,
  literate=,
  showstringspaces=false,
  columns=fullflexible,
  keepspaces=true,
}

\lstdefinestyle{oacontext}{
  basicstyle=\rmfamily\footnotesize,
  keywords={}, keywords=[2]{},
  literate=,
  showstringspaces=false,
  breaklines=true,
  columns=fullflexible,
  keepspaces=true,
}

\definecolor{promptbg}{rgb}{0.96,0.96,0.96}  
\lstdefinestyle{oaprompt}{
  basicstyle=\ttfamily\footnotesize,
  backgroundcolor=\color{promptbg},
  frame=single,
  framesep=4pt,
  rulecolor=\color{black!30},
  xleftmargin=4pt,
  xrightmargin=4pt,
  keywords={}, keywords=[2]{},
  literate={—}{{\textemdash}}1,
  showstringspaces=false,
  breaklines=true,
  columns=fullflexible,
  keepspaces=true,
}

\definecolor{rahi}{rgb}{1.00,0.91,0.62}  
\newcommand{\rahl}[1]{{\setlength{\fboxsep}{1pt}%
  \colorbox{rahi}{\ttfamily\bfseries\color{oaext}#1}}}

\newcommand{\oasubrule}{\par\vspace{1pt}{\color{black!30}\hrule height 0.4pt}\vspace{1pt}}

\graphicspath{{figures/}}

\usepackage{hyperref}
\hypersetup{colorlinks=true,linkcolor=blue,citecolor=blue,urlcolor=blue}

\newcommand{\OA}{\textsc{Object Aligner}}
\newcommand{\score}{\ensuremath{\mathrm{s}}}            
\newcommand{\sleaf}{\ensuremath{\sigma}}                
\newcommand{\given}{\,;\,}
\DeclareMathOperator*{\argmax}{arg\,max}
\newcommand{\refmap}{\ensuremath{\pi}}                  

\newcommand{\cmark}{\ensuremath{\checkmark}}            
\newcommand{\xmark}{\ensuremath{\times}}                
\newcommand{\pmark}{(\ensuremath{\checkmark})}          

\def\BibTeX{{\rm B\kern-.05em{\sc i\kern-.025em b}\kern-.08em
    T\kern-.1667em\lower.7ex\hbox{E}\kern-.125emX}}

\definecolor{editmark}{rgb}{0.00,0.30,0.80}

\let\PARstart\IEEEPARstart

\begin{document}

\bstctlcite{IEEEexample:BSTcontrol}

\title{Object Aligner: A Configurable JSON Schema Similarity Score for Graphs, Applied to LLM Prompt Optimization}

\author{Jan~Drchal%
  \thanks{J.~Drchal is with the Artificial Intelligence Center, Faculty of
    Electrical Engineering, Czech Technical University in Prague, Czech
    Republic (e-mail: drchajan@fel.cvut.cz).}%
  \thanks{This work was created with the state support of the Technology
    Agency of the Czech Republic within the Sigma Programme, Project
    No.~TQ01000100.}%
  \thanks{Preprint. This is a submitted version of a manuscript under review
    at IEEE Access; it has not been peer reviewed.}}

\markboth
{Drchal: Object Aligner: A Configurable JSON Schema Similarity Score for Graphs}
{Drchal: Object Aligner: A Configurable JSON Schema Similarity Score for Graphs}

\maketitle

\begin{abstract}
Large language models (LLMs) are often asked to produce JSON conforming to a fixed
schema, powering information extraction, tool calling,
agentic planning, and knowledge-graph construction. Measuring how closely an output matches a gold
reference is essential yet surprisingly hard: exact match is brittle, text similarity
ignores structure, and an LLM judge is expensive, opaque, and non-deterministic.
We address this with \emph{\OA{}} (OA), an open-source Python library
that scores two JSON objects deterministically by recursively aligning their
trees (the Hungarian algorithm for unordered collections, sequence alignment for
ordered ones) and awarding partial credit at the granularity the schema declares.
The \OA{} is configured entirely through a set of JSON Schema extensions, so
adapting it to a new task involves annotating a schema rather than writing code.
Complex structured data, however, are rarely flat trees: records may form
graphs or hypergraphs keyed by arbitrary identifiers, breaking the assumptions of prior
similarity metrics.
Our central contribution, \emph{referential
alignment}, closes this gap by inferring a bijection between gold and candidate
identifiers and scoring every reference through it, so the score is invariant to
relabeling. Since recovering this bijection exactly is graph isomorphism, the \OA{}
approximates it with Weisfeiler--Leman color refinement. An
order-sensitive \emph{sequence} regime targets ranking and planning.  
Since the same alignment localizes every mismatch, the \OA{} emits ranked repair
suggestions at no extra cost. Used as a reward inside the GEPA
prompt optimizer, \OA{} helps or stays neutral across all datasets.
\end{abstract}


\section{Introduction}
\label{sec:introduction}

\PARstart{C}{omparing} complex structured data, i.e., measuring how closely two structured
objects agree, is a basic operation in many settings: automated testing, change detection
between versions, record linkage, and the evaluation of machine-learning systems against
reference outputs. It has become more urgent with large
language models (LLMs). Along with free-form
text, an LLM is often asked to return data conforming to a fixed schema (most often
JSON) so that its output can be parsed, stored, and consumed directly by downstream
systems: information extraction, document understanding, tool and function calling,
agentic planning, and knowledge-graph construction all rely on it. JSON is among the most
widely used data formats in practice, and although it is essentially a tree, cross-references
between records allow it to encode general graphs and hypergraph structures
(Fig.~\ref{fig:oa-example}). Judging whether two such objects agree (for instance a model prediction against a
gold reference) is therefore both common and hard: the comparison must look past
cosmetic differences such as reordering or identifier renumbering and credit what is
structurally correct.

\begin{figure*}[t]
\centering
\begin{lstlisting}[style=oajsonschema, basicstyle=\ttfamily\scriptsize]
schema = {"type":"object", "properties": {
  "people": {"type":"array", "order":"align",
   "items": {"type":"object", "keyImportance":0.0, "valueImportance":1.0, "properties": {
    "id":   {"type":"integer", "idScope":"person"},
    "name": {"type":"string", "score":"exact", "valueWeight":2.0},
    "role": {"type":"string", "score":"exact", "enum":["Manager", "Engineer", "Intern"]}}}},
  "mentorships": {"type":"array", "order":"align", "ignoreExcess":true,
   "items": {"type":"object", "properties": {
    "mentor": {"type":"integer", "ref":"person"}, "mentee": {"type":"integer", "ref":"person"}}}},
  "agenda": {"type":"array", "order":"fixed", "items": {"type":"string", "score":"jaro", "threshold":0.3}},
  "period": {"type":"array", "prefixWeights":[1, 1],
   "prefixItems": [{"type":"integer", "score":"invdiff"}, {"type":"integer", "score":"invdiff"}]}}}
\end{lstlisting}
\vspace{1pt}
\begin{minipage}[t]{0.49\textwidth}
\begin{lstlisting}[style=oajsondata, basicstyle=\ttfamily\scriptsize]
gold = {
 "people": [
  {"id": 1, "name": "Alice", "role": "Manager"},
  {"id": 2, "name": "Bob",   "role": "Engineer"},
  {"id": 3, "name": "Carol", "role": "Engineer"},
  {"id": 4, "name": "Dave",  "role": "Engineer"},
  {"id": 5, "name": "Dave",  "role": "Engineer"},
  {"id": 6, "name": "Eve",   "role": "Intern"}],
 "mentorships": [
  {"mentor": 1, "mentee": 4},
  {"mentor": 2, "mentee": 3},
  {"mentor": 3, "mentee": 5}],
 "agenda": ["intro", "reviews", "closing"],
 "period": [2023, 2025]}
\end{lstlisting}
\end{minipage}\hfill
\begin{minipage}[t]{0.49\textwidth}
\begin{lstlisting}[style=oajsondata, basicstyle=\ttfamily\scriptsize]
candidate = {
 "people": [
  {"id": 90, "name": "Dave",  "role": "Engineer"},
  {"id": 91, "name": "Alice", "role": "Manager"},
  {"id": 92, "name": "Dave",  "role": "Engineer"},
  {"id": 93, "name": "Carol", "role": "Engineer"},
  {"id": 94, "name": "Bob",   "role": "Engineer"}],
 "mentorships": [
  {"mentor": 91, "mentee": 92},
  {"mentor": 94, "mentee": 93},
  {"mentor": 93, "mentee": 90}],
 "agenda": ["reviews", "intro", "closin"],
 "period": [2023, 2026]}
\end{lstlisting}
\end{minipage}
\caption{Motivating example, used as the running example throughout. The schema
(top) declares \texttt{people} ids and \texttt{mentorships} that reference them;
\OA{} extensions are in \textcolor{oaext}{\textbf{orange}}
(Appendix~\ref{app:keywords}). \texttt{gold} and \texttt{candidate} encode the
\emph{same} org chart even though \texttt{candidate} renumbers every id and reorders
the people list. The two \texttt{Dave}/\texttt{Engineer} records are
property-identical \emph{twins}: only graph structure tells them apart
(Section~\ref{sec:oa-ref}). The candidate also omits the intern \texttt{Eve},
transposes the first two \texttt{agenda} items and drops a character in the
third, and is a year off on \texttt{period}. Under the identifier bijection
$\refmap=\{1\!\mapsto\!91,\,2\!\mapsto\!94,\,3\!\mapsto\!93,\,
4\!\mapsto\!92,\,5\!\mapsto\!90,\,6\!\mapsto\!\bot\}$ every \texttt{mentorship}
matches; the residual errors yield $\score=0.77$.}
\label{fig:oa-example}
\end{figure*}

When the scoring procedure itself must be controlled---reproducible and auditable---LLM-based judging is not a solution. 
LLM-as-a-judge~\cite{geval}
replaces a human evaluator with another model, but it inherits the very difficulty we are
trying to escape: the judge must itself be prompted, calibrated, and validated; it is
costly, noisy, and non-deterministic; it is prone to position and verbosity bias; and its
verdicts are opaque, which makes them hard to audit. Generic text-similarity
metrics~\cite{bertscore, bleurt, comet} avoid these costs but discard the structure
entirely. What is needed
instead is a deterministic, schema-aware score that compares a prediction directly against a
gold object and credits the parts it gets right.

We introduce \emph{Object Aligner} (OA), a highly configurable similarity score\footnote{Following 
common usage in the evaluation literature, we use the terms ``score'' and ``metric'' interchangeably, 
even though \OA{} is not a metric in the strict mathematical sense.}
for structured outputs, computed by a recursive, schema-driven alignment of gold and
candidate trees that matches each node's children with the appropriate optimal-alignment
algorithm: the Hungarian algorithm for unordered collections and a sequence-alignment dynamic
program for ordered ones (such as our example in Fig.~\ref{fig:oa-example}). By design, the
score is deterministic, decomposable, and schema-aware, so partially correct objects receive
partial credit at the granularity declared by the schema. The comparison is driven entirely by
a schema expressed as a small set of JSON Schema extensions (per-field weights, leaf
comparators, and ordering and reference semantics) so adapting \OA{} to a new task amounts
to annotating a schema rather than writing code. \OA{} ships as an open-source Python
module.

Such a score has many uses; this study concentrates on one of them, prompt
optimization. Coaxing an LLM to
produce the \emph{correct} structured output is difficult: performance is highly sensitive to
how the task is phrased, which examples are shown, and how the schema and its constraints are
explained. Therefore building a reliable pipeline is increasingly an exercise in prompt engineering.
Prompt-optimization (PO) frameworks automate this search by iteratively proposing and
refining prompts, and can match or surpass careful manual tuning at a fraction of the human
effort~\cite{aposurvey}.

Regardless of the method, the reward that \emph{measures} candidate quality determines what
the PO search can achieve: no search can optimize for quality its reward cannot see. That
reward is computed over a labeled dataset and evaluated for many candidate prompts across
the loop. This is exactly where the \OA{} fits---a
deterministic, decomposable structural reward---though prompt optimization is far from its
only use. The common alternative of using an LLM to score candidates or to reflect on their
failures is computationally demanding and non-deterministic, expensive to run at the scale a
search requires, hard to reproduce and audit, and sensitive to prompt and position
bias---the same problems that motivate a controlled score in the first place.

The core idea---side-by-side recursive Hungarian matching of gold and candidate
trees---was described concurrently by STED~\cite{sted} and ExtractBench~\cite{extractbench}
and implemented in Stickler~\cite{stickler}. 
Our publicly recorded
implementation\footnote{\url{https://github.com/aic-factcheck/prompt_opt}, committed
2024-12-20.} predates all of them, and we treat this part of the work as a simultaneous
discovery. Taking this shared idea as our starting point, we chose to extend it; hence our
main contributions are:

\begin{itemize}
  \item An open-source Python library implementing
    \OA{}\footnote{\url{https://github.com/aic-factcheck/object_aligner}}, configured
    entirely through JSON Schema extensions and deployable as a drop-in reward for existing
    prompt-optimization frameworks such as DSPy~\cite{dspy}, GEPA~\cite{gepa}, and
    TextGrad~\cite{textgrad}.
  \item The application of a deterministic structural score as the reward signal of a prompt
    optimizer. It is the first PO pipeline driven by schema-aware partial credit
    over nested structures, to our knowledge.
  \item \emph{Referential alignment} for (hyper)graphs: A scoring regime invariant to the
    relabeling of identifiers, which infers a bijection between gold and candidate
    identifiers and scores every reference through it, approximated with
    Weisfeiler--Leman color refinement~\cite{weisfeiler1968reduction}
    since recovering it exactly is graph isomorphism (Section~\ref{sec:oa-ref}).
  \item \emph{Per-list sequence semantics}: A per-list choice between order-agnostic
    matching, a monotone, insertion/deletion-aware regime suited to ranking and planning
    tasks, and \emph{positional tuples}---fixed-arity sequences whose slots carry
    position-specific meaning, each with its own importance and comparator
    (Section~\ref{sec:oa-lists}).
  \item \emph{Deterministic ranked feedback} for prompt optimization: The same gold--candidate
    alignment that produces the score also pinpoints where the candidate departs from the
    gold, and emits these mismatches as repair operations---edits that would bring the
    candidate closer to the gold---ranked by the exact amount of score each recovers, so the
    optimizer is pointed at the changes that matter most, with no LLM call
    (Section~\ref{sec:oa-feedback}).
  \item An empirical study on both synthetic data, which allows us to isolate and control the specific
    properties of the score, and real-world datasets.\footnote{The code for all experiments is available at \url{https://github.com/aic-factcheck/object_aligner_paper}, ensuring the reproducibility.}
\end{itemize}

The remainder of the paper is organized as follows. Section~\ref{sec:related} reviews
the structural similarity metrics, concurrent structured-output scorers, and prompt
optimization. Section~\ref{sec:object-aligner} defines the \OA{} and its extensions.
Section~\ref{sec:datasets} describes the datasets, and Section~\ref{sec:results} reports the
experiments. Section~\ref{sec:discussion} discusses the findings,
Section~\ref{sec:limitations} the limitations, and Sections~\ref{sec:conclusion}
and~\ref{sec:future-work} conclude and outline future work.

\section{Related Work}
\label{sec:related}

\OA{} builds on a long line of work on comparing structured objects and is
motivated by a recent one: prompt optimizers that consume cheap, reliable
reward signals. We review structural similarity first, then the concurrent
structured-output scorers, and close with the prompt-optimization frameworks
that define the use case.

\subsection{Similarity Scoring for Structured Data}
\label{sec:rel-metrics}

\paragraph{Tree edit distance.}
Comparing trees is classically cast as tree edit distance (TED): the
minimum-cost script of node insertions, deletions, and relabelings that turns
one tree into the other~\cite{tai1979tree}. For ordered trees the
Zhang--Shasha algorithm computes the TED in polynomial
time~\cite{zhang1989simple}, with APTED being the current state of the
art~\cite{pawlik2016apted}. For unordered
trees, the problem is NP-hard, and Zhang's \emph{constrained} edit distance
restores tractability by requiring disjoint subtrees to map to disjoint
subtrees, which reduces the matching of each node's children to a bipartite
assignment~\cite{zhang1996constrained}: the closest classical relative of
\OA{}'s recursion. 
X-Diff applies the same restriction to unordered XML change
detection~\cite{xdiff2003}. All of these return an edit cost under uniform,
hand-set operation costs: they have no notion of a schema, per-field
importance, or graded leaf similarity.

\paragraph{Optimal assignment as an evaluation primitive.}
The Hungarian algorithm~\cite{kuhn1955hungarian} has a long history in
evaluation. In coreference resolution, CEAF scores a system by
maximum-weight bipartite matching between gold and predicted entity
clusters~\cite{luo2005ceaf}, in contrast to link-based
MUC~\cite{vilain1995muc} and LEA~\cite{moosavi2016lea}, and 
mention-based B$^3$~\cite{bagga1998b3}. Template-filling
evaluation extends CEAF by one level of recursion---templates are matched on
the aggregate similarity of their slot fillers~\cite{grit,iterx}---but
remains specific to flat MUC-style templates. In computer vision, DETR uses
the same matching as training loss for set prediction~\cite{detr}. Most
directly, the bipartite approximation of graph edit distance reduces an
NP-hard comparison to a single linear assignment over node edit
costs~\cite{riesen2009bipartite}.
\OA{} applies similar idea recursively to schema-typed trees, with the cost matrix computed bottom-up
from configurable leaf comparators.

\paragraph{Semantic-graph metrics.}
Invariance to identifier renumbering---the property motivating referential
alignment (Section~\ref{sec:oa-ref})---has been studied mainly for semantic
graphs. Smatch scores two Abstract Meaning Representation (AMR) graphs by 
searching for a one-to-one alignment of their variables with a restart-dependent hill-climbing
heuristic~\cite{smatch}. 
S2match replaces binary concept equality with graded
embedding similarity~\cite{s2match}, and Smatch++ solves the alignment
optimally as an integer linear program~\cite{smatchpp}. WWLK instead compares
AMR graphs through Weisfeiler--Leman (WL) node embeddings and optimal
transport~\cite{wwlk}. 
Discrete shared WL colors underlie the
optimal-assignment graph kernel WL-OA~\cite{kriege2016wloa}, the closest prior
use of joint refinement for matching vertices (cf.\
Section~\ref{sec:oa-ref}). These metrics achieve exactly the renumbering
invariance \OA{} targets, but for a single formalism: the Smatch family
operates on a flat triple set, none recurses into nested values, and none
exposes schema-level configuration. We make this relationship concrete by
evaluating \OA{} on AMR (Section~\ref{sec:dataset-amrbio}), where Smatch's
variable alignment is the same problem that \OA{} solves as referential matching
from a schema.

\paragraph{Practical comparison of JSON and records.}
JSON diff libraries such as DeepDiff recurse through nested objects and can
ignore list order, but they pair list items greedily rather than by optimal
assignment and expose only an unweighted distance with no field importance or
graded leaf similarity~\cite{deepdiff}.
Probabilistic record linkage is a mature precedent for \emph{declarative
per-field comparison}: Fellegi--Sunter-based tools such as Splink attach fuzzy
comparators and weights to each field of a flat
record~\cite{fellegi1969theory,splink}.
\OA{} generalizes this declarative pattern to trees of arbitrary depth. The idea
of identifying records through their references rather than their attributes is 
rooted in the relational model.
Chen's entity--relationship model represents a relationship as a relation keyed by the primary keys of the entities
it links and, when attributes alone cannot identify a record, resolves it
recursively through its relationships with other records~\cite{chen1976er}---the
reference-driven disambiguation that referential alignment performs up to
identifier relabeling (Section~\ref{sec:oa-ref}). For LLM
outputs specifically, function-calling benchmarks check predicted calls by
exact AST matching, recursing into argument values for type and equality
but offering no graded leaf similarity, field importance, or alignment of records
by reference~\cite{bfcl}, while the flexible
alternative, LLM-as-a-judge~\cite{geval}, reintroduces the cost, noise, and
calibration burden discussed in Section~\ref{sec:introduction}. Recent surveys
likewise call for a more rigorous, reproducible LLM evaluation
methodology~\cite{sheikhi2026beyond}.

\subsection{Concurrent Structured-Output Scorers}
\label{sec:rel-concurrent}

To our knowledge, three systems developed concurrently with the \OA{}
(Section~\ref{sec:introduction}) target the same problem.
Stickler~\cite{stickler}, an AWS Labs library (without an accompanying paper),
scores LLM outputs against Pydantic models with per-field comparators and
weights; lists of structured models are Hungarian-matched, but pairs scoring
below a match threshold are not recursed into and are counted as false
detections in a confusion-matrix report aimed at document-processing audits.
STED~\cite{sted} computes a tree edit distance with Hungarian child matching
at every level and embedding-based leaf costs, controlled by a handful of
global weights with no schema or per-field configuration, targeting
the consistency benchmarking of repeated generations. 
ExtractBench~\cite{extractbench} is primarily a benchmark; its
evaluator reads a per-field scoring configuration from an annotated schema but
aligns arrays with an LLM call rather than an assignment solver.

Table~\ref{tab:related} summarizes these differences, with Smatch++ included as
the closest non-JSON relative. Several axes have no full counterpart in
the three concurrent systems. First, none is invariant to identifier
renumbering: renumbering the identifiers of Fig.~\ref{fig:oa-example}
changes their score even though the encoded graph is unchanged, whereas
referential alignment (Section~\ref{sec:oa-ref}) scores every reference through
an inferred identifier bijection: among these systems, only Smatch++
offers the same guarantee (but limited to AMR graphs). Second,
none lets the schema declare, per list, whether order is part of the answer:
STED matches children order-agnostically throughout, Stickler always
Hungarian-matches structured lists, and ExtractBench delegates ordering to the
judge LLM, whereas \OA{}'s \texttt{order} keyword selects per sequence between
the order-agnostic (optimal bipartite) and order-sensitive (monotone,
insertion/deletion-aware) regimes (Section~\ref{sec:oa-lists}). Relatedly,
none treats a fixed-arity sequence as a \emph{positional tuple}---slots whose
meaning differs by position, each with its own importance and comparator,
aligned one-to-one by position (Section~\ref{sec:oa-lists}): STED and
ExtractBench match such arrays interchangeably like any other list. Third, only
\OA{} emits optimizer-ready feedback: repair operations ranked by exact score
deltas derived from the match tree (Section~\ref{sec:oa-feedback});
Stickler's reports target human auditors, and STED returns a bare scalar.
Finally, two design choices make the \OA{} suitable for the inner loop of prompt
optimization, its motivating use case. The score is continuous, with no
threshold gating (unlike Stickler), and scoring requires no embedding or LLM
calls (unlike STED and ExtractBench); rollouts can thus be scored
deterministically and reproducibly, without the per-call model inference those
scorers require.

\begin{table}[t]
\centering
\caption{Design-axis comparison of \OA{} with the concurrent structured-output
scorers and the closest semantic-graph metric. \pmark{} denotes partial
support; ``dependency-free'' means no embedding or LLM calls at scoring time;
superscript letters refer to the notes below.}
\label{tab:related}
\footnotesize
\setlength{\tabcolsep}{3.5pt}
\begin{tabular}{@{}p{4.1cm}ccccc@{}}
\toprule
 & \rotatebox{90}{\OA{} (ours)}
 & \rotatebox{90}{Stickler~\cite{stickler}}
 & \rotatebox{90}{STED~\cite{sted}}
 & \rotatebox{90}{ExtractBench~\cite{extractbench}}
 & \rotatebox{90}{Smatch++~\cite{smatchpp}} \\
\midrule
Schema-declared comparators \& weights & \cmark & \pmark$^{a}$ & \xmark & \cmark & \xmark \\
Recursive optimal assignment           & \cmark & \pmark$^{b}$ & \cmark & \xmark$^{c}$ & \pmark$^{d}$ \\
Deterministic, dependency-free scoring & \cmark & \pmark$^{e}$ & \xmark & \pmark$^{e}$ & \cmark \\
Continuous score (no threshold gating) & \cmark & \xmark & \cmark & \cmark & \cmark \\
Per-list order semantics               & \cmark & \xmark & \xmark & \xmark & \xmark \\
Tuples (per-slot weights/comparators) & \cmark & \xmark & \xmark & \xmark & \pmark$^{h}$ \\
Identifier-renumbering invariance      & \cmark & \xmark & \xmark & \xmark & \cmark$^{f}$ \\
Optimizer feedback (repair ops, deltas)& \cmark & \pmark$^{g}$ & \xmark & \xmark & \xmark \\
\bottomrule
\end{tabular}
\par\smallskip
\begin{minipage}{\columnwidth}\raggedright\scriptsize
$^{a}$Pydantic-first; JSON Schema via vendor-prefixed extensions.
$^{b}$Hungarian matching for lists of structured models only; recursion gated
by a match threshold.
$^{c}$Array alignment via the LLM call.
$^{d}$Optimal ILP alignment over a flat triple set; not recursive.
$^{e}$Deterministic comparators available; embedding/LLM comparators common.
$^{f}$Via AMR variable alignment; formalism-specific.
$^{g}$HTML reports intended for human audit.
$^{h}$Triples are fixed $\langle$source,\,relation,\,target$\rangle$ slots,
matched whole once a one-to-one node alignment is chosen; Smatch++ adds a
per-triple weight and one graded \emph{concept}-slot comparator, but no
configurable comparator or weight per position.
\end{minipage}
\end{table}

\subsection{Prompt Optimization}
\label{sec:rel-promptopt}

A prominent line of automatic prompt optimization uses LLMs as optimizers: APE
generates instruction candidates and keeps the best-scoring~\cite{ape}, 
OPRO iteratively
proposes prompts conditioned on the trajectory scored so far~\cite{opro},
evolutionary variants maintain a mutating
population~\cite{promptbreeder,evoprompt}, and error-driven refinement
clusters failure cases into prompt updates~\cite{pandy2026error}. DSPy makes
the reward interface
explicit: a multi-stage LM program is compiled against an arbitrary
user-supplied metric~\cite{dspy}, which its MIPROv2 optimizer maximizes by
Bayesian optimization over instructions and few-shot
demonstrations~\cite{miprov2}. TextGrad instead propagates natural-language
``gradients''---textual critiques of intermediate outputs---through compound
systems~\cite{textgrad}.
Across all of these frameworks, the metric is supplied by the user. Where prompt
optimizers have been applied to structured-output tasks, the reward has been
exact-match triple F1 for knowledge-graph construction~\cite{kgcapo},
exact-match parse accuracy for semantic parsing~\cite{sammo}, or, at most, a
graded span-overlap measure for flat entity extraction~\cite{toxhabits};
graded \emph{structural} rewards such as Smatch or graph edit distance have
appeared only as reinforcement-learning signals for weight
tuning~\cite{naseem-etal-2019-rewarding,structllm}. To the best of our knowledge, no published
prompt-optimization pipeline has driven the search with schema-driven partial
credit over nested structures.

We adopt GEPA~\cite{gepa} as the optimizer in all the experiments for two
reasons. First, it is sample-efficient: candidate mutations are screened on
minibatches rather than full validation sweeps, and new candidates are drawn
from a Pareto frontier over individual training instances, preserving
diversity. 
The authors report that it outperforms MIPROv2 and matches or beats
GRPO-style reinforcement learning at a fraction of the rollouts. Second, it is
reflective: like TextGrad and other feedback-driven optimizers, it
updates prompts from natural-language critique, but here that critique is
returned by the metric itself: each mutation is proposed by an LLM that reads
execution traces together with \emph{textual feedback} from the metric, so
its metric interface is a (score, feedback) pair rather than a bare
scalar. \OA{} exactly fills this interface: a deterministic score for Pareto
bookkeeping and structurally faithful repair feedback for reflection
(Section~\ref{sec:oa-feedback}).

\section{Object Aligner}
\label{sec:object-aligner}

This section provides a description of the \OA{}. We first fix the data
model and the two-phase scoring it uses (Section~\ref{sec:oa-overview}), and then
instantiate it for the primitive, sequence, and map nodes
(Sections~\ref{sec:oa-leaves}--\ref{sec:oa-objects}).
We then develop three extensions that are the focus of this
paper: order-sensitive sequence alignment with insertion/deletion support
(Section~\ref{sec:oa-lists}), referential alignment for (hyper)graphs, which
compares structured outputs up to identifier renumbering
(Section~\ref{sec:oa-ref}), and deterministic, optimizer-shaped
feedback that turns the score into a learning signal
(Section~\ref{sec:oa-feedback}). We close with the formal properties and
cost of the procedure (Section~\ref{sec:oa-properties}). 

\subsection{Overview and data model}
\label{sec:oa-overview}

The \OA{} takes a \emph{gold} object $g$, a \emph{candidate} (predicted) object $p$, and a
\emph{schema} $S$, and returns a similarity \emph{score}
$\score(g, p \given S) \in [0, 1]$, where $1$ denotes a perfect match and $0$ a
complete mismatch. By design, the
score is (i) \emph{deterministic}---the same inputs always yield the same
number\footnote{User-defined primitive type comparators, such as semantic embedding-based scores, may break determinism.}; 
(ii) \emph{schema-aware}---the comparison is
driven entirely by $S$, so partially correct objects receive partial credit at
the granularity the schema declares; and (iii) \emph{decomposable}---the
top-level score is an explicit weighted aggregate of child scores, which
makes feedback possible (Section~\ref{sec:oa-feedback}).

We model both $g$ and $p$ as value trees. A value is either a \emph{primitive}
(a string, a number, or a boolean), the empty value $\bot$, a \emph{sequence}
(an ordered tuple of values), or a \emph{map} (a JSON \texttt{object}, a finite set of key--value pairs
with distinct keys). The schema $S$ is an annotated tree of the same shape: each
node fixes the expected type and the local scoring choices---which primitive
comparator to use, how to weight children, whether a sequence is order-sensitive, and
whether a primitive acts as an identifier. 
Our implementation represents the value trees as JSONs, whereas the schema takes the form of 
JSON Schema extensions documented in Appendix~\ref{app:keywords}.

The alignment descends $g$, $p$, and $S$
in lockstep, producing a score at every node and aggregating upward to the root.
Fig.~\ref{fig:oa-example} shows a concrete schema for the running example, a
small organization chart: six \texttt{people} in one identifier scope, directed
\texttt{mentorships} between them, an order-sensitive \texttt{agenda}, and a
\texttt{period} year tuple. The schema also exercises several \OA{}-specific
extensions beyond the plain JSON Schema (e.g., \texttt{keyImportance},
\texttt{valueWeight}, \texttt{enum}, \texttt{prefixItems}, and fuzzy
\texttt{score} comparators).

The scoring at every internal node proceeds in two phases. In the \emph{alignment}
phase, \OA{} fixes a correspondence between the children of $g$ and those of
$p$---a partial matching that says which gold child is compared against which
candidate child. In the \emph{scoring} phase, it aggregates the per-pair child
scores over the fixed correspondence into a single number in~$[0,1]$.

The cost that drives the matching depends on the node type and schema configuration. 
For \emph{sequences}, the elements are matched by either the Hungarian algorithm or the order-sensitive dynamic programming approach.
For a \emph{map}, matching involves only the Hungarian algorithm over \emph{key similarity}. 
Moreover, referential alignment (Section~\ref{sec:oa-ref}) introduces identifiers and references, analogous to
primary and foreign keys in relational databases.
Primitive types (leaves) are scored directly. A node where
one or both sides are the empty value $\bot$ is also scored directly: two empties
agree (score $1$), whereas a value present on only one side scores the schema's
\texttt{nullScore} (default $0$). Algorithm~\ref{alg:align} states the dispatch, and 
the following subsections describe each case.

\begin{algorithm}[t]
\caption{Recursive alignment and scoring $\score(g,p\given S)$}
\label{alg:align}
\begin{algorithmic}[1]
\REQUIRE gold value $g$, candidate $p$, schema node $S$
\ENSURE score in $[0,1]$
\IF{$S$ is an identifier or reference field}
  \RETURN symbolic id/reference score \COMMENT{Section~\ref{sec:oa-ref}}
\ELSIF{$g=\bot$ or $p=\bot$}
  \RETURN $1$ if both empty, else \texttt{nullScore} (default $0$)
\ELSIF{$S$ is a primitive}
  \RETURN $\sleaf(g,p)$
\ELSIF{$S$ is a sequence}
  \STATE \textbf{align} children by $M_{ij}=\score(g_i,p_j)$:
  \STATE \quad order-agnostic (Hungarian) or order-sensitive (DP)
  \RETURN $\tfrac1D\sum_{(i,j)\in\mathcal M} M_{ij}$ \COMMENT{Eq.~\eqref{eq:reorder}}
\ELSIF{$S$ is a map}
  \STATE \textbf{align} keys by $K_{ij}=\kappa(k_i,k'_j)$ \COMMENT{keys only}
  \STATE \textbf{score} matched pairs: $v_{ij}\gets\score(g[k_i],p[k'_j])$
  \RETURN weighted aggregate of $\{K_{ij}\}$, $\{v_{ij}\}$ \COMMENT{Eq.~\eqref{eq:dict}}
\ENDIF
\end{algorithmic}
\end{algorithm}

\subsection{Primitive types}
\label{sec:oa-leaves}

A primitive node carries a similarity function $\sleaf:\mathcal V\times\mathcal
V\to[0,1]$ on primitive values $\mathcal V$. For strings, $\sleaf$ is a configurable
comparator: exact match $\mathbbm 1[g=p]$, a normalized edit- or
sequence-distance similarity, such as Jaro~\cite{jaro1989advances} or
Levenshtein~\cite{levenshtein1966binary}. For numbers, we may choose an exact match, inverse difference ($1/(1+|g-p|)$); Booleans are
compared exactly. Any primitive type can be scored using a user-supplied comparator function (e.g., semantic embedding similarity for text).

\subsection{Sequences}
\label{sec:oa-lists}

A sequence node compares two tuples $g_1,\dots,g_n$ and $p_1,\dots,p_m$ whose
elements share a child schema. The per-pair matching cost is the full recursive
score $M_{ij}=\score(g_i,p_j)$, so whatever similarity the children
carry (nested maps, further sequences) propagates into the correspondence. 
The Object Aligner implements two regimes: 

\paragraph{Order-agnostic.}
When the element order is irrelevant (a set of entities), the alignment is the
\emph{optimal} bipartite matching: pad $M$ to a square $d\times d$ matrix
($d=\max(n,m)$) with zero rows/columns for absent elements and choose
\begin{equation}
  \refmap^\star = \argmax_{\refmap}
    \sum_{i=1}^{d} M_{i,\refmap(i)},
  \label{eq:assignment}
\end{equation}
the maximum-weight assignment, solved optimally using the Hungarian
algorithm~\cite{kuhn1955hungarian}. Writing
$\mathcal{M}=\{(i,\refmap^\star(i)) : M_{i,\refmap^\star(i)}>0\}$ for the
genuinely matched pairs, the score is
\begin{equation}
  \score(g,p\given S) = \frac{1}{D}\sum_{(i,j)\in\mathcal M} M_{ij}.
  \label{eq:reorder}
\end{equation}
An unmatched gold element is \emph{missing}, and an unmatched candidate element is
\emph{excess}; each occupies a slot scoring $0$. The denominator
$D = |\mathcal M| + n_{\mathrm{miss}} + n_{\mathrm{exc}}$ counts the matched pairs
plus the penalized unmatched elements, and whether the missing or excess elements count
toward $D$ is schema-configurable.

\paragraph{Order-sensitive.}
When the order is itself part of the answer (steps of a plan, ranked items),
admissible matchings are restricted to the \emph{monotone} ones, which preserve
the left-to-right order while allowing insertions and deletions. The optimal monotone
matching is computed by the standard sequence-alignment dynamic program and scored
under the same denominator convention as in Eq.~\eqref{eq:reorder}. Unlike the
order-agnostic regime, a transposition breaks the monotone path and is penalized
on both elements.

\paragraph{Tuples and prefixes.}
The two regimes above treat a sequence as unbounded. In practice, we often work with 
\emph{tuples}: a fixed number of positional slots whose meaning differs by
position (a coordinate triple, a labeled header), so each slot deserves its own
importance rather than being matched interchangeably. We unify both as a
\emph{prefix}: a fixed-arity positional head followed by an optional
variable-length tail. The prefix elements are aligned one-to-one by position;
with per-position importances $v_1,\dots,v_a$ ($a$ the prefix arity, uniform by default) their scores are 
combined as
\begin{equation}
  s_{\mathrm{pre}} = \frac{\sum_{k=1}^{a} v_k\, \score(g_k, p_k\given S)}
                          {\sum_{k=1}^{a} v_k},
  \label{eq:prefix-score}
\end{equation}
whereas the tail is aligned by one of the two regimes above into $s_{\mathrm{rest}}$
(a pure tuple is just the empty-tail case). The two blend as
\begin{equation}
  \score(g,p\given S) = \frac{w_p\, s_{\mathrm{pre}} + w_r\, s_{\mathrm{rest}}}
                              {w_p + w_r},
  \label{eq:prefix}
\end{equation}
with prefix and tail importances $w_p,w_r$.

\subsection{Maps}
\label{sec:oa-objects}

The map is aligned \emph{by its keys alone}. Given gold keys $k_1,\dots,k_n$ and
candidate keys $k'_1,\dots,k'_m$, the alignment is optimal bipartite matching
(Eq.~\eqref{eq:assignment}) on the key-similarity matrix $K_{ij} = \kappa(k_i, k'_j)$
where $\kappa$ is a key comparator (exact, fuzzy, or user-defined). 
Note that the \emph{values are not consulted when choosing which keys pair}: maps
are matched by their labels only. This is deliberate. The keys already identify
what each entry represents, so they suffice to decide what is compared to what,
and matching values would force the user to weight keys against values: two
scales with no obvious, schema-independent trade-off. Keys fix the correspondence, 
the values then grade it.
If the key-only rule is too limiting, the collection can be modelled as an
order-agnostic sequence of $\langle\text{key},\text{value}\rangle$ tuples (Section~\ref{sec:oa-lists}),
recovering the relational view of a map as a matched set of typed pairs. 

Given the matched key pairs
$\mathcal{M}=\{(i,\refmap^\star(i)) : K_{i,\refmap^\star(i)}>0\}$, the
\emph{scoring} phase recurses on the paired values,
$v_{ij}=\score(g[k_i],p[k'_j])$ (unmatched keys contribute $0$), and aggregates a
key term and a value term:
\begin{align}
  s_{\mathrm{key}} &= \tfrac{1}{d}\textstyle\sum_{(i,j)\in\mathcal M} K_{ij}, \\
  s_{\mathrm{val}} &= \frac{\sum_{(i,j)\in\mathcal M} \omega_i\, v_{ij}}
                            {\sum_{i=1}^{d} \omega_i}, \\
  \score(g,p\given S) &= \frac{w_K\, s_{\mathrm{key}} + w_V\, s_{\mathrm{val}}}
                              {w_K + w_V},
  \label{eq:dict}
\end{align}
with key weight $w_K$, value weight $w_V$, and per-property weights $\omega_i$.
The \OA{}'s default $w_K=0$ (schema keyword \texttt{keyImportance}) treats keys as fixed \emph{scaffolding} therefore, 
only the values are scored.
Setting $w_K>0$ suits open-vocabulary
extraction, where the model also \emph{chooses} the keys and obtaining a key right
is itself part of the task.

\subsection{Referential alignment}
\label{sec:oa-ref}

Many structured outputs are really \emph{(hyper)graphs}: some records carry an
identifier at which other parts point back. The identifier values are
meaningless; two correct extractions may number their records differently
(Fig.~\ref{fig:oa-example}) and a primitive-by-primitive comparison would
punish this as an error. We want a score \emph{invariant to relabeling of identifiers}. 
Referential alignment does this: the schema marks one primitive field as an
\emph{identifier} and others as \emph{references} to it, and \OA{} infers a
bijection between gold and candidate identifiers and then scores every reference
through it.

\paragraph{Scopes, records, and references.}
The schema keywords are \texttt{idScope} and \texttt{ref}
(Fig.~\ref{fig:oa-example}), which behave like primary
and foreign keys. An \texttt{idScope} on a primitive inside a sequence names a
\emph{scope} whose items are its \emph{records}; a \texttt{ref} of the same type
points to a named scope from anywhere. In the example, the \texttt{person} scope
has six records (the \texttt{people}, keyed by \texttt{id}), and each
\texttt{mentorship} references two of them (\texttt{mentor} and \texttt{mentee}).
Identifiers and references are matched \emph{symbolically} through the bijection,
not by raw values, so their primitive comparator is ignored.

\paragraph{Deriving the bijection.}
To recover the bijection for a scope, we align its gold and candidate records as
an order-agnostic sequence (Eq.~\eqref{eq:reorder}): writing $d_i$ for the
gold records and $d'_j$ for the candidate ones, form the record cost matrix
$C_{ij}=\score(d_i, d'_j)$, solve the assignment (Eq.~\eqref{eq:assignment}), and
read the bijection $\refmap:\text{gold ids}\to\text{candidate ids}$ off the optimal
pairs (partial if the sides differ in size---unmatched gold ids map to $\bot$ and
references to them score $0$; excess candidate ids stay outside the bijection's
image, so any reference pointing at one cannot match its gold counterpart and
scores $0$). Deriving the bijection involves two steps.
First, we \emph{mask} the scope's own id field: its value is arbitrary, so
comparing it would only penalize the right pairing. Instead, we treat it as a
perfect match (score $1$), so it drops out of the comparison, and records are
paired by everything \emph{except} their id---in the example, by name and role,
which already pair each gold person with the right candidate. Second, we never compare a reference by value,
while the scope it points
to is still unresolved, as that would use a bijection we have not yet computed. 
Thus, we resolve scopes in dependency order (topological sort, lexicographic tie-break
for determinism). 
A reference into the \emph{same} scope, and any cycle of mutually
referential scopes, has no such order, so there we warn and fall back to
property-only alignment.

\begin{figure*}[t]\centering
\setlength{\tabcolsep}{5pt}\renewcommand{\arraystretch}{0.5}
\resizebox{\textwidth}{!}{%
\begin{tabular}{ccc!{\vrule width 0.6pt}ccc}
\multicolumn{3}{c}{\textbf{gold}} & \multicolumn{3}{c}{\textbf{candidate}}\\[1pt]
{round 0} & {round 1} & {round 2} &
{round 0} & {round 1} & {round 2}\\
\includegraphics[height=2.9cm]{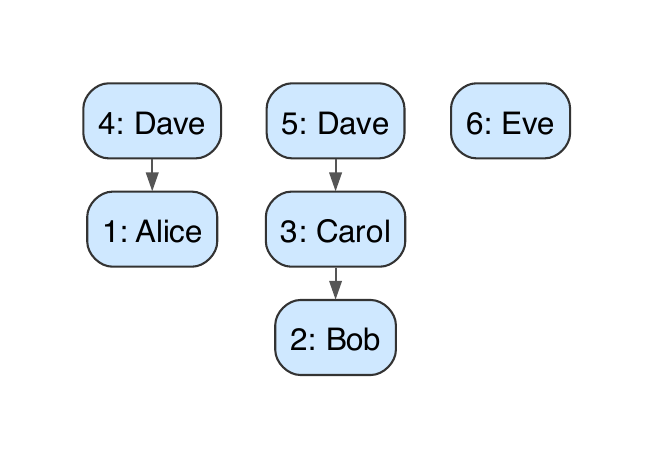} &
\includegraphics[height=2.9cm]{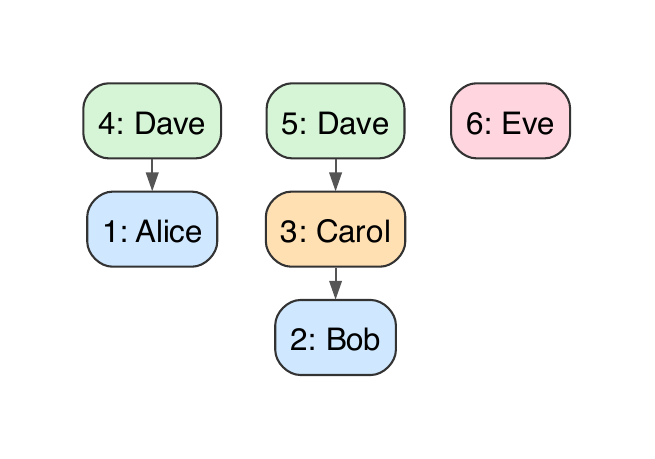} &
\includegraphics[height=2.9cm]{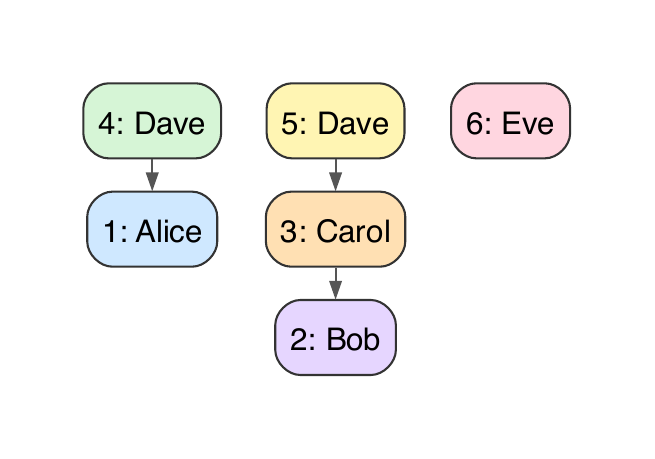} &
\includegraphics[height=2.9cm]{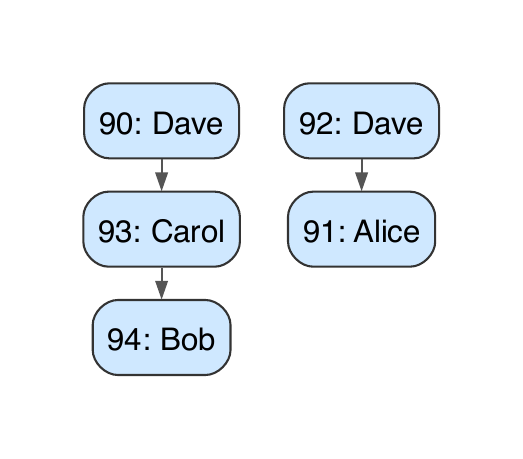} &
\includegraphics[height=2.9cm]{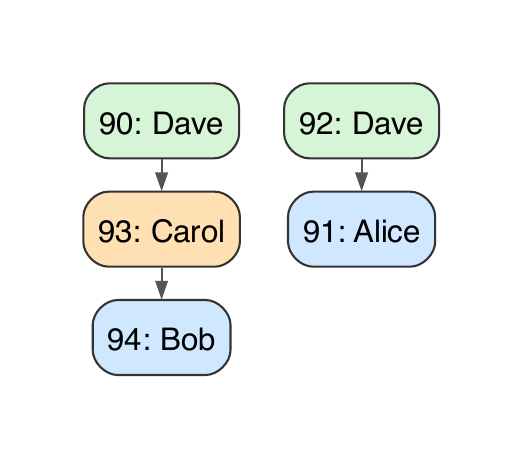} &
\includegraphics[height=2.9cm]{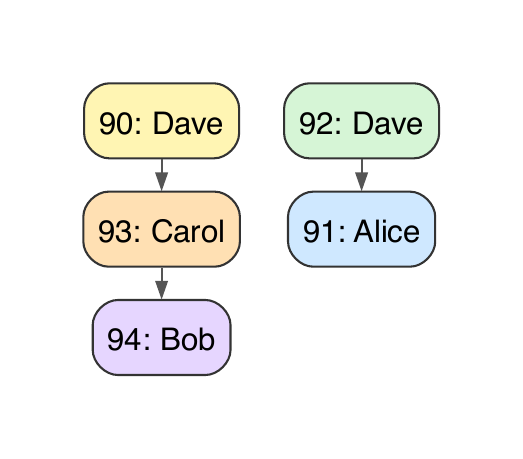}\\
\end{tabular}}
\caption{Joint \emph{tie-break} 1-WL refinement over the
disjoint union $\mathcal G_g\sqcup\mathcal G_c$ (\texttt{gold} left,
\texttt{candidate} right; rounds left-to-right, the last is stable); a color change
marks a refinement step. The relabeling is shared, so equal colors are comparable
across the two sides: by the final round every \texttt{candidate} record carries the
color of its \texttt{gold} counterpart, which pins the
\texttt{Dave}/\texttt{Engineer} twin bijection.}
\label{fig:oa-wl}
\end{figure*}

\paragraph{Breaking ties by structure.}
Masking ids can leave the assignment underdetermined: when two records agree on
all their \emph{other} attributes (the two \texttt{Dave}/\texttt{Engineer}
\emph{twins} in the example) their rows in $C$ are identical and attributes give
no reason to prefer one pairing. The twins may still differ in \emph{structure}:
one Dave is mentored by \texttt{Alice}, and the other by \texttt{Carol}
(Fig.~\ref{fig:oa-wl}). Resolving this exactly is again graph isomorphism, far
costlier than the cubic Hungarian assignment that we already pay.
We therefore use a near-linear $1$-dimensional Weisfeiler--Leman (1-WL) color
refinement~\cite{weisfeiler1968reduction,shervashidze2011wlkernels} as a tractable
approximation to that test. It gives each node a color and repeatedly refreshes it from its neighbors' colors until
structurally distinct nodes are separated.

The colors must be comparable \emph{across} the two sides, and refining each graph
on its own does not give that: 1-WL numbers colors in the order signatures appear,
so gold's color~$3$ and candidate's color~$3$ are unrelated. We therefore refine
\emph{jointly}: one graph per side---a node per record, and a directed edge for
each reference linking two records, labeled by the scalar fields of the
reference-bearing record (its \emph{carrier}); in the example each
\texttt{mentorship} is a single \texttt{mentor}\,$\to$\,\texttt{mentee} edge between
two \texttt{person} nodes---over their \emph{disjoint union} $\mathcal G_g\sqcup\mathcal G_c$ (gold,
candidate; Fig.~\ref{fig:oa-wl}), each round
recomputing a vertex's signature from its own color and the sorted multiset of its
incident-edge descriptors,
\begin{equation}
  \mathrm{sig}^{t+1}(v) = \big(\, c^{t}(v),\ \mathrm{sort}\,\{\!\!\{\,
    (\delta, f, \ell, c^{t}(u)) \,\}\!\!\} \,\big),
  \label{eq:wl-refresh}
\end{equation}
where the multiset ranges over all incident edges $(v,u,f,\ell)\in\mathcal E$ in
both directions $\delta\in\{\text{in},\text{out}\}$, each contributing its
direction $\delta$ (for a mentorship, whether $v$ is the \texttt{mentor} or the
\texttt{mentee} end), the relationship's label $f$ (here
\texttt{mentor}\,$\to$\,\texttt{mentee}), an edge label $\ell$, and the neighbor's
color $c^{t}(u)$. The label $\ell$ is compared by \emph{exact} equality, so it
carries only what is already settled on both sides---the carrier's own scalar
fields together with any of its references into already-resolved scopes---while
fuzzy attributes\footnote{Scalars compared by a graded similarity rather 
than exact equality (e.g., Jaro string distance or inverse numeric difference), 
whose partial scores have no place in WL's exact-equality color refinement.} 
and references into the still-unresolved current scope are left
out. In the running example, a \texttt{mentorship} holds nothing but its two
references, so $\ell$ is empty and the twin \texttt{Dave}s separate only through
their neighbors' colors. Were each link to also record a \texttt{since}
year---say \texttt{Alice}\,$\to$\,\texttt{Dave} since~\texttt{2019} and
\texttt{Carol}\,$\to$\,\texttt{Dave} since~\texttt{2021}---their two incoming edges
would carry $\ell=\texttt{2019}$ versus $\ell=\texttt{2021}$, splitting the
look-alike \texttt{Dave}s by this exact-equal scalar in a single round, without the
extra refinement their mentors' differing colors otherwise require.

A single \emph{intern} table shared by both
sides (\textsc{Relabel}, Algorithm~\ref{alg:wl}) maps each round's signatures to
integer colors $c^{t+1}(v)$, so identical signatures get the same color on gold and
candidate---comparable by construction, with no seed, anchor, or learned embedding,
and without ever consulting the bijection we derive (no circularity). This
joint relabeling is the standard Weisfeiler--Leman construction, \emph{concordant}
across graphs by design~\cite{shervashidze2011wlkernels}. 
The closest prior method
to use such discrete shared colors for \emph{matching} vertices is the
optimal-assignment kernel WL-OA~\cite{kriege2016wloa}. We depart from it in two
ways: our graphs carry directed, relation- and scalar-labeled edges with references into
already-resolved scopes, and the colors enter only as an infinitesimal tie-break
(below) rather than as the assignment objective itself. In Fig.~\ref{fig:oa-wl}, our
construction pins the twins, and in the final round, each \texttt{candidate} Dave carries
the color of its gold match.

In the default \emph{tie-break} mode, the colors enter the cost matrix only as an
infinitesimal bonus $\epsilon\,b_{ij}$, with $b_{ij}=\mathbbm 1[\,\text{colors
agree}\,]$ and $\epsilon$ below the smallest gap between distinct attribute costs; therefore,
the structure breaks \emph{only} exact ties and never overrides what the attributes
decide. \OA{} also implements an optional \emph{blend} mode that seeds the
refinement from each record's attributes and mixes structure into the cost at a
fixed weight rather than as a tie-break, which suits cases where attributes already
separate most records. We omit its details and experiments for brevity.

\begin{algorithm}[t]
\caption{\textsc{Relabel}: intern signatures to comparable integer colors}
\label{alg:wl}
\begin{algorithmic}[1]
\REQUIRE a signature $\mathrm{sig}[v]$ for every vertex $v\in V$ of the union
         $\mathcal G_g\sqcup\mathcal G_c$
\ENSURE one integer color per vertex, comparable across both sides
\STATE $D \gets \textsc{Sort}\big(\{\,\mathrm{sig}[v] : v\in V\,\}\big)$
       \COMMENT{distinct signatures, canonical order}
\STATE $\mathrm{color}\gets$ empty map
\FOR{$i = 0,\dots,|D|-1$}
  \STATE $\mathrm{color}[D_i] \gets i$ \COMMENT{next free integer, reused for repeats}
\ENDFOR
\RETURN $\{\, v \mapsto \mathrm{color}[\,\mathrm{sig}[v]\,] : v\in V\,\}$
\end{algorithmic}
\end{algorithm}

\paragraph{Scoring references.}
With the bijection in hand, the main alignment pass (Algorithm~\ref{alg:align})
treats identifier fields as labels worth $1$ and scores a reference as correct if and only if
it points, through the inferred mapping, at the same record the gold reference does
(and that target exists among the candidate ids; otherwise it scores $0$). In
Fig.~\ref{fig:oa-example} the masked name/role comparison, with the twins split by
structure, recovers
$\refmap=\{1\!\mapsto\!91,2\!\mapsto\!94,3\!\mapsto\!93,4\!\mapsto\!92,
5\!\mapsto\!90,6\!\mapsto\!\bot\}$, under which every \texttt{mentorship} matches
exactly despite every id differing; the object still loses credit for the omitted
\texttt{Eve} and the perturbed \texttt{agenda}/\texttt{period}, for a final
$\score=0.77$.

\subsection{Optimizer feedback}
\label{sec:oa-feedback}

A scalar reward tells an optimizer \emph{how} well a prompt does, but not
\emph{what} to change. Because \OA{}'s score is decomposable, we can answer the
second question deterministically and \emph{without an LLM}: the same alignment
that produced the score also localizes the deficit, which we render as a compact,
prescriptive feedback string suitable for the natural-language reflection slot of a
reflective prompt optimizer such as GEPA~\cite{gepa} or TextGrad~\cite{textgrad}---in
DSPy~\cite{dspy}, for instance, GEPA is exposed as an optimizer whose metric returns
this textual feedback alongside the scalar score. This
contrasts sharply with LLM-as-judge feedback~\cite{geval}, which is noisy,
non-deterministic, and must be prompted and calibrated.

Walking the match tree, every imperfect leaf or structural mismatch becomes a
\emph{repair operation} with an associated \emph{score delta}---the amount of
total score recovered by fixing it. For a child contributing with effective
weight $c_w$ and value score $s_w$, the delta is
$\Delta(\mathrm{op}) = c_w\,(1 - s_w)$, and, under the fixed assignment chosen by
alignment, summing these deltas over the full set $\mathcal{O}$ of repair operations
accounts for the entire score deficit exactly,
$\sum_{\mathrm{op}\in\mathcal{O}} \Delta(\mathrm{op}) = 1 - \score(g,p\given S)$,
that is, the gap between a perfect score and the one achieved.
Operations are typed (primitive replacement, missing/extraneous key, fuzzy key
rename, missing/excess sequence element, wrong reference, whole-subtree
replacement, etc.) and each carries a path to its location.

Operations are ranked by $\Delta$, the top $K$ are kept (those below a minimum
delta are dropped), and each is rendered through a per-type template. 
A trailing \emph{synthesis} line names the dominant error category when one type accounts
for most of the displayed deficit. The per-type templates are configurable.
For the running example of Fig.~\ref{fig:oa-example} (scored at $0.77$), the
feedback rendered by the default templates is shown in Fig.~\ref{fig:oa-feedback}.
Each line is a deterministic template substitution, and the deltas are derived from
the exact decomposition described above.

\begin{figure*}[t]
\centering
\fbox{\begin{minipage}{0.98\textwidth}
\fontsize{7.5}{9}\selectfont\ttfamily\noindent
The prediction scored 0.77 (deficit 0.23). Top 5 of 5 fix locations:\\
1. /agenda/0: extraneous list item 'reviews'. Removing it recovers +0.062.\\
2. /agenda/2: missing list item 'reviews'. Adding it recovers +0.062.\\
3. /period/1: expected 2025, got 2026. Fixing this recovers +0.062.\\
4. /people: list is missing item \{'id': 6, 'name': 'Eve', 'role': 'Intern'\}. Adding it recovers +0.042.\\
5. /agenda/3: expected 'closing', got 'closin'. Fixing this recovers +0.003.\\
The deficit is spread across multiple issue types (missing-list-item, primitive-value, extraneous-list-item).
\end{minipage}}
\caption{Deterministic GEPA-style optimizer feedback for the running
example (Fig.~\ref{fig:oa-example}, $\score=0.77$): fix locations ranked by score
delta, each rendered from a per-type template, closing with the synthesis line.
The reported $\Delta$ values come straight from the deficit decomposition, so the
feedback is faithful to the score by construction.}
\label{fig:oa-feedback}
\end{figure*}

\subsection{Properties and cost}
\label{sec:oa-properties}

\OA{} is \emph{deterministic} and reproducible: every choice point (assignment,
WL relabeling, tie-breaks) is resolved by a canonical, seed-independent rule. The
score is \emph{bounded} in $[0,1]$ and awards partial credit at schema
granularity. It is \emph{decomposable}, which precisely enables the
attribution and feedback of Section~\ref{sec:oa-feedback}.

The cost is dominated by per-node work. A sequence or map alignment of arity
$d$ solves one assignment problem, $O(d^3)$, over an $O(d^2)$-entry matrix whose
\emph{sequence} entries are recursive alignments (a \emph{map}
instead scores cheap key comparisons, recursing only on the $\le d$ matched
pairs); order-sensitive sequences replace the assignment with an $O(nm)$ dynamic
program. Referential alignment adds one assignment per scope plus WL refinement,
which costs $O\!\big(R\,(|V|+|E|)\big)$ with $R\le|V|$ rounds over the scope's
reference graph. Summed over the tree, the cost is dominated by the assignments,
each cubic in its collection's local arity $d$, where $d$ is the number of
matched siblings---fixed and small for an object's keys, so only list lengths
grow with the document. A single such list is already $O(n^3)$, and nesting
growing lists multiplies this further, but in practice these lists stay short and
shallow, so the cost remains a non-issue.

%
\section{Datasets}
\label{sec:datasets}
The following datasets are used in Section~\ref{sec:results} to demonstrate the Object Aligner properties,
focusing mainly on Referential Alignment (RA) and order semantics.
The list contains both synthetic (Org2Graph and Facts2Order) and real-world data.
We briefly introduce each dataset; worked examples, schemas, and native task metrics are deferred to the per-dataset appendices (Appendix~\ref{app:datasets}, native metrics in Appendix~\ref{app:native}). 
Real-world datasets were converted to JSON under an OA schema and, in some cases, reframed for our needs.

\subsection{Org2Graph}
\label{sec:dataset-org2graph}

In natural information extraction corpora, the difficulty of recovering
\emph{referential structures} is tangled up with parsing, domain vocabulary,
and annotation noise, so they cannot isolate \emph{why} referential alignment
helps or fails. Org2Graph (O2G) is a synthetic generator that converts this difficulty
into controlled variables. Each instance is a templated
natural-language paragraph describing a fictional organization, and the task is
to extract the graph it describes as JSON with five scopes: \texttt{people}
and \texttt{companies} are records carrying identifiers, and
\texttt{employment}, \texttt{acquaintance}, and \texttt{partnership} records
link them through references, foreign-key style. Records and links carry one
categorical property each (\texttt{title}, \texttt{industry}, \texttt{role},
and the two relation types). Because the paragraph is rendered from the gold
graph (every link surfaces as exactly one sentence) extraction is
unambiguous by construction. Fig.~\ref{fig:intrinsic-ra-sample}
(Appendix~\ref{app:intrinsic}) shows a small instance including gold JSON
output.

Three parameters control where the difficulty lies; each scope draws its
categorical values from a fixed \emph{codebook} (its vocabulary of allowed
entries, or\emph{codes}). \emph{Twin density} sets the fraction of records that are
\emph{property-twins}---identical in every scored attribute, thus telling them
apart, and hence routing references correctly, requires the reference
structure itself. Within a scope, exactly one set of twins shares identical 
scored attributes, whereas every other record maintains a distinct combination of them. \emph{Value obfuscation} replaces the
readable codebook entries with opaque codes under a fixed legend never shown
to the model (the paragraph says ``engineer,'' the gold value is
\texttt{T07}), so an optimizer can recover the vocabulary only by aggregating
evidence across examples. \emph{Vocabulary width} sets each scope's codebook
size ($24/20/20/12/12$ vs.\ $6/6/6/3/3$); narrow codebooks make
property-twins frequent.

\subsection{Facts2Order}
\label{sec:dataset-facts2order}

Validating \emph{order} sensitivity requires a task whose correct order is
unique, stated explicitly in the input, and inexpensive to corrupt by a measured
amount. Facts2Order (F2O) is a
synthetic generator built to satisfy these properties. Each instance lists $N$ short
sentences, each stating one sortable key for a named entity (an integer
quantity, a date, or an ordinal word). The task is to output the index
permutation that sorts the items by their key
(Fig.~\ref{fig:facts2order-sample}). The generator exposes a
few independent parameters, and the study instantiates several configurations that
differ in their settings. The \emph{number of items} $N$ fixes the length of
the permutation to recover; we use different values of $N$ across experiments
(exact ranges in Section~\ref{sec:intrinsic} and Section~\ref{sec:results-order}).
The \emph{key type} is an integer quantity, a date, or an ordinal word. Each
instance may carry a variable number of \emph{irrelevant distractor sentences}
that state no sortable key, and merely pad the input. Finally, the \emph{sort
key} is presented in one of two regimes: \emph{stated}, where the key is the
lone numeric clause on each item, or \emph{hidden}, where each item additionally
carries $2$--$3$ \emph{numeric decoy} clauses on other attributes (e.g., a price
in dollars, a length in centimeters) whose values are unconstrained, with the
numeric clauses shuffled per item, so the key is never identifiable as the lone
number and the model must discover \emph{which} measure to sort on.
Keys are drawn distinctly, so the gold
order is unique, while the distractor sentences and decoy clauses leave
it untouched.
Because gold and candidate are index permutations, disorder is exactly
measurable via the Kendall distance~\cite{kendall1938rank} (the minimum number of adjacent
transpositions between two orderings), so corruptions can be dosed.

\subsection{SciERC}
\label{sec:dataset-scierc}

SciERC~\cite{scierc} annotates $500$ computer-science abstracts with
scientific entities, coreference clusters, and typed relations, which is our
natural-text extraction benchmark, where difficulty arises from parsing and
domain vocabulary. The target output is a
\emph{graph}: \texttt{entities} as nodes joined by reference-linked
\texttt{relations}, extracted from the abstract as JSON. 
A worked instance is in
Fig.~\ref{fig:scierc-sample} (Appendix~\ref{app:dataset-scierc}). 
A \emph{simplified} form is used. 
Because reproducing exact character offsets is poorly suited to the
text-generation nature of decoder-only LLMs~\cite{genner}, we drop all span
offsets and keep only surface text: each coreference cluster becomes one entity
whose \texttt{mentions} array holds its surface forms (its closed-vocabulary
\texttt{type}, one of six, by majority vote).
The raw annotations carry no entity identifiers, so we assign each entity
a synthetic \texttt{id} (\texttt{e0}, \texttt{e1}, \ldots) in the cluster
first-occurrence order purely to link the relation endpoints.
The original mention-pair
relations are projected to the cluster level (one of the seven closed
\texttt{predicate}s), discarding any relation whose endpoints fall outside a
cluster. We use original splits with no
subsampling ($349/50/100$ train/val/test).

\subsection{BioRED}
\label{sec:dataset-biored}

BioRED~\cite{biored} is a document-level biomedical relation-extraction
corpus of PubMed abstracts preprocessed into the same graph output as SciERC
(Section~\ref{sec:dataset-scierc})---coreference-cluster \texttt{entities} joined by
reference-linked \texttt{relations}. 
Its distinctive feature is that each gold entity id is an opaque
normalized concept accession not derivable from the abstract, so relations must, 
in principle, be routed by referential alignment rather than literal-id matching.
We draw a seeded $50/50/100$ train/val/test split, with train and val
subsampled from BioRED's train/dev pools, and the test split is the full BioRED test
split.
See details in Appendix~\ref{app:dataset-biored}.

\subsection{Bio AMR}
\label{sec:dataset-amrbio}

The Bio AMR Corpus~\cite{bioamr} annotates biomedical sentences
(cancer-related PubMed text) with Abstract Meaning Representation~\cite{amr}
graphs: each sentence maps to a rooted, directed graph of concept nodes and
labeled edges. 
Like BioRED (Section~\ref{sec:dataset-biored}), AMR's variable names are
arbitrary and hidden, so referential alignment alone can carry out relation
routing. Bio AMR is the purest case: the same concept recurs on many nodes, so
content individuates nothing, and structure does all the work.
We convert each gold PENMAN graph into a JSON
object of \texttt{nodes} and \texttt{relations},
dropping the source variable letters so that the model emits its own arbitrary node
ids and only the graph structure is scored; a worked instance is shown in
Fig.~\ref{fig:amrbio-sample} (Appendix~\ref{app:dataset-amrbio}).
The registered schema scores this referentially and pairs with literal-id \emph{strict}
ablation (Appendix~\ref{app:dataset-amrbio}), which measures how much of the score
referential alignment carries. We draw a seeded, unstratified $100/100/200$
train/val/test subsample of the corpus.

\subsection{NATURAL PLAN (Trip Planning)}
\label{sec:dataset-natural-plan}
This dataset and the one that follows target OA's \emph{order} semantics.
NATURAL PLAN~\cite{naturalplan} is a natural-language planning benchmark. 
We use its \emph{Trip Planning} task, which asks for a
strictly \emph{ordered} itinerary---a sequence of cities to visit, each for a
given number of days, reachable only over a stated set of direct flights. Unlike the synthetic
Facts2Order (Section~\ref{sec:dataset-facts2order}), the gold order here is imposed
by a genuine planning constraint rather than designed in, so a correct plan must
obtain both the set of stays \emph{and} their sequence right. Each instance's input
\texttt{context} is based on the day budget, the
per-city stays and meeting windows, and the direct-flight list; the gold output
is taken straight from the released \texttt{cities}/\texttt{durations} fields (Fig.~\ref{fig:natural-plan-sample},
Appendix~\ref{app:dataset-natural-plan}). We draw a difficulty-stratified
$100/100/200$ train/val/test split keyed on the number of cities ($3$--$10$).

\subsection{ROCStories}
\label{sec:dataset-rocstories}

ROCStories~\cite{rocstories} is a corpus of five-sentence everyday
commonsense stories. The correct order is fixed by narrative
coherence in real text rather than designed in by sortable keys, so recovering it
demands commonsense reasoning. We retain only stories
that segment into exactly $N=5$ sentences; the input \texttt{context} presents those
sentences scrambled and labeled $1..N$, and the gold output is the index permutation
that restores the original reading order, emitted as a JSON \texttt{index} list
(Fig.~\ref{fig:rocstories-sample}, Appendix~\ref{app:dataset-rocstories}). The data
key is deliberately neutral (\texttt{indices}, not \texttt{order}), so neither schema
nor seed prompt hints that the task is reordering. We draw a $100/100/200$ train/val/test split (all instances are $N=5$). 
See more details in Appendix~\ref{app:dataset-rocstories}.

\section{Results}
\label{sec:results}

We evaluate the \OA{} in two stages. An \emph{intrinsic} study
(Section~\ref{sec:intrinsic}) validates its two novel mechanisms with no LLM in
the loop, in the perturbation-based design of STED~\cite{sted}: the score
must stay high under benign corruptions of the gold and fall, in proportion,
under harmful ones. 
Because STED already validates the machinery, the two
metrics share (Hungarian-style order-invariant matching above all) our
experiments target only \OA{}'s additions: referential alignment (RA,
Section~\ref{sec:oa-ref}) and the order-sensitive \emph{sequence} regime
(Section~\ref{sec:oa-lists}). An \emph{extrinsic} study
(Section~\ref{sec:results-setup}--\ref{sec:results-feedback}) measures
what these mechanisms buy as the reward and
feedback signal driving prompt optimization (GEPA~\cite{gepa}).

\subsection{Intrinsic validation of the alignment mechanisms}
\label{sec:intrinsic}

The setup is simple: a deterministic generator produces gold outputs,
controlled corruptions damage them, and we measure how the score
$\score(g,p\given S)$ responds (AUROC, rank correlations). A good score must
do two things at once: stay constant under changes that preserve meaning and
drop in proportion to genuine damage. RA must ignore identifier relabeling yet 
register every structural edit, and
the sequence regime must grade reorderings yet degrade gracefully when items
are dropped or inserted.

\paragraph{Invariance to relabeling.}
Referential alignment promises that the score does not depend on how a
candidate numbers its records: identifiers are matched through the inferred
bijection $\refmap$, not by value (Section~\ref{sec:oa-ref}). We generate $100$
gold Org2Graph graphs (Section~\ref{sec:dataset-org2graph}) in the
plain-narrow configuration, with sizes, edge densities, and twin density
$t\sim U[0,1]$ drawn at random (Algorithm~\ref{alg:intrinsic-ra-equiv},
Appendix~\ref{app:intrinsic}), and for each emit five exact copies under
independent identifier relabelings ($500$ pairs). Each copy is
graph-isomorphic to its gold, so a relabel-invariant score must give all five
exactly $1$; the per-gold variance is the diagnostic. RA passes
exactly (score $1.000$, variance $0$ on all $100$ golds) even on the nine
where \emph{every} record is a property-twin and the bijection is pinned only
by the reference structure (see Fig.~\ref{fig:intrinsic-ra-sample}, Appendix~\ref{app:intrinsic}). The
\textbf{plain} ablation (\texttt{idScope}/\texttt{ref} dropped, identifiers
compared by value---our stand-in for any reference-unaware metric) averages
$0.738$ with per-gold variance up to $5.5\times10^{-3}$: a flawless output
forfeits a quarter of its credit, by an amount that depends on the arbitrary
choice of identifiers---exactly the noise a score should not inject.

\begin{figure*}[t]
\centering
\includegraphics[width=0.92\textwidth]{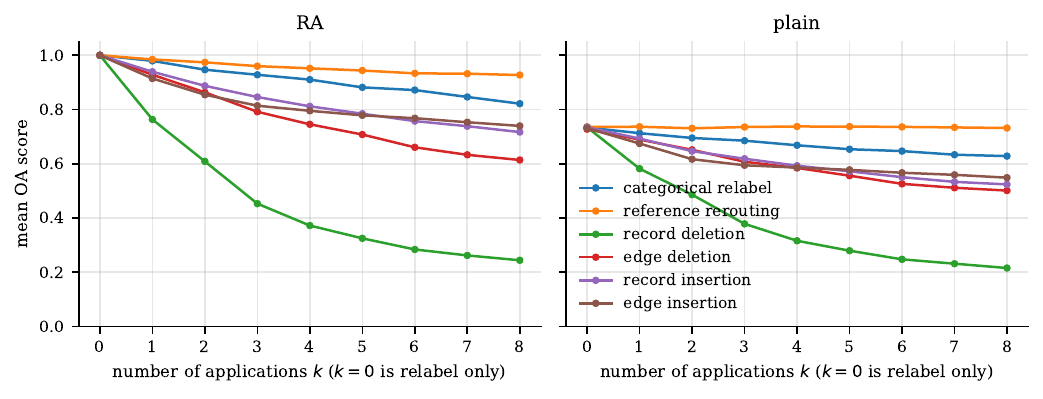}
\caption{Sensitivity to damage (Org2Graph generator), pooled over
graph sizes and seeds: each of the six edit types applied \emph{alone}, $k$
times, on top of an identifier relabel ($k{=}0$ is the relabel only).
\emph{Left}: RA; \emph{right}: the plain ablation. RA starts at $1.0$ and
degrades monotonically under every operation; plain starts at $0.74$---the
relabel alone already costs it---and is blind to reference rerouting (flat
orange curve), the one edit that corrupts routing without touching any
record's attributes.}
\label{fig:intrinsic-ra-perop-pooled}
\end{figure*}

\paragraph{Sensitivity to damage.}
Invariance alone is trivial; therefore, the second experiment checks that RA still
falls, in proportion, under genuine damage. On $100$ fresh golds we apply a
\emph{single} edit type $k{=}0,\dots,8$ times on top of a relabeled copy
($k{=}0$ is the relabel alone): recoding a categorical value, rerouting a
reference, deleting or inserting a record, or an edge---$5400$ pairs
(Algorithm~\ref{alg:intrinsic-ra-perop}, Appendix~\ref{app:intrinsic}). 
RA starts at $1.0$ and decreases
monotonically with $k$ under every operation
(Fig.~\ref{fig:intrinsic-ra-perop-pooled}, left; per-operation Spearman
$-0.66$ to $-0.89$, Table~\ref{tab:intrinsic-ra-spearman} in
Appendix~\ref{app:intrinsic}), and separates damaged from clean almost
perfectly (AUROC $0.996$); the residual is correct, not noise---$42$ of the
$4800$ edited candidates, mostly reroutes between property-twins, are
graph-isomorphic to their gold and correctly scored $1$. Plain separates significantly
worse (AUROC $0.835$), and \emph{reference rerouting} shows why
(Fig.~\ref{fig:intrinsic-ra-perop-pooled}, right): the one edit that corrupts
the graph without touching any record's attributes moves only RA
($1.00\!\to\!0.93$ over eight reroutes; Spearman $-0.660$), while plain remains
flat ($0.74\!\to\!0.73$; $-0.011$)---the damage most specific to graph
extraction is precisely the damage it cannot see.
A control sweep without the relabel confirms that this gap is specific to
relabeled identifiers, not the edits themselves: with identifiers matching by
value, RA and plain nearly coinciding except under reference rerouting and
categorical relabel (data omitted for brevity).

\paragraph{Sequence regime.}
The sequence regime (Section~\ref{sec:oa-lists}) claims to \emph{grade}
order: a nearly-sorted output must outscore a scrambled one. On
\textsc{Facts2Order} (Section~\ref{sec:dataset-facts2order})---configured
with the item count drawn per instance as $N\sim U\{3..12\}$, the key type
drawn uniformly from \{integer, date, ordinal\}, $0$--$3$ irrelevant distractor
sentences, and the \emph{stated}-key regime (no numeric decoys). 
OA scores the
index permutations directly, so the surface sentences are ignored. We corrupt $300$
gold orderings in two ways. The first is pure reorderings: adjacent transpositions up to a target
Kendall distance, block reversals, and moves. The second is length-changing
edits of the next paragraph. Together these yield $4639$ pairs (generation:
Algorithm~\ref{alg:intrinsic-order}; worked instance:
Fig.~\ref{fig:intrinsic-order-sample}). 
Each pair is scored in four ways: the
\textbf{sequence} alignment under test; the order-agnostic \textbf{set}
alignment (Hungarian matching of the label bag), standing in for STED-style
matching; \textbf{exact} match (Perfect-Match Ratio); and
Kendall~$\tau$, the graded reference one would hand-pick for this
task ($0$ on any non-permutation). 
On pure reordering the sequence score falls smoothly with Kendall distance
(see Table~\ref{tab:intrinsic-order}; Spearman $-0.706$), while separating sorted
from corrupted perfectly (AUROC $1.000$). 
The exact match also separates perfectly---it is $0$ on every corrupted candidate by construction---but
offers no gradient. The set
score is \emph{exactly constant} at $1.0$ for every pure reordering (the
label bag never changes): zero order signal, with AUROC $0.705$ owing to 
incidental length changes alone. Kendall~$\tau$ is graded, as designed
(Spearman $-0.621$); OA's sequence alignment tracks this task-specific
reference without being built for the task. Fig.~\ref{fig:intrinsic-order-partial} 
(Appendix~\ref{app:intrinsic}) shows this behavior graphically.

\begin{figure}[t]
\centering
\footnotesize
\begin{verbatim}
Here are the items:
Item 1: Saiph weighs 37 kilograms.
Item 2: Antares weighs 696 kilograms.
Item 3: Arcturus weighs 933 kilograms.
Item 4: Bellatrix weighs 883 kilograms.

gold: {"indices": [1, 2, 4, 3]}
      (mass order: 37 < 696 < 883 < 933)
\end{verbatim}
\normalsize
\smallskip
\footnotesize
\setlength{\tabcolsep}{3pt}
\begin{tabular}{llcccc}
\toprule
candidate & \texttt{indices} & sequence & set & exact & $\tau$ \\
\midrule
gold (no edit)    & \texttt{[1,2,4,3]} & $1.000$ & $1.000$ & $1$ & $+1.000$ \\
one adjacent swap & \texttt{[1,4,2,3]} & $0.600$ & $1.000$ & $0$ & $+0.667$ \\
fully reversed    & \texttt{[3,4,2,1]} & $0.143$ & $1.000$ & $0$ & $-1.000$ \\
one item dropped  & \texttt{[1,2,4]}   & $0.750$ & $0.750$ & $0$ & $0.000$ \\
\bottomrule
\end{tabular}
\caption{A \textsc{Facts2Order} instance ($N{=}4$, integer key) with
its unique gold order, and four candidates scored four ways---the
per-instance view of Table~\ref{tab:intrinsic-order}. Only the
\emph{sequence} score grades both the reorderings and the dropped item.}
\label{fig:intrinsic-order-sample}
\end{figure}

Models also drop or add new records; therefore, we further delete or
insert $k{=}1..3$ items per candidate (insertions duplicate existing ones).
A single dropped item sends both the exact match and Kendall~$\tau$ to $0$:
Kendall~$\tau$ is undefined between orderings of different item sets (and scores
$0$), and the exact match trivially fails on any
length change. Instead the OA's sequence
alignment loses credit in proportion to the edit count through its
gap-aligned denominator (Section~\ref{sec:oa-lists}).
Under these pure length-changing edits (no accompanying reordering), 
the set score coincides with the sequence. Nevertheless, across both
axes, the sequence is the only scorer that stays graded
(Table~\ref{tab:intrinsic-order}), using the same generic, schema-driven
machinery that scores every other structure in the \OA{}.

\begin{table}[t]
\centering
\caption{Mean score by corruption family (\textsc{Facts2Order} generator, $4639$ pairs). Adjacent transpositions at Kendall distance $d$ keep the
items and change only their order; deletions and insertions change the
length.}
\label{tab:intrinsic-order}
\footnotesize
\setlength{\tabcolsep}{5pt}

\begin{tabular}{l cccc}
\toprule
Corruption & sequence & set & exact & Kendall $\tau$ \\
\midrule
none ($d{=}0$, gold) & $1.000$ & $1.000$ & $1.000$ & $1.000$ \\
\addlinespace[2pt]
\multicolumn{5}{l}{\emph{adjacent transpositions, Kendall distance $d$}}\\
\quad $d{=}1$ & $0.737$ & $1.000$ & $0.000$ & $0.852$ \\
\quad $d{=}2$ & $0.641$ & $1.000$ & $0.000$ & $0.751$ \\
\quad $d{=}3$ & $0.581$ & $1.000$ & $0.000$ & $0.772$ \\
\quad $d{=}4$ & $0.538$ & $1.000$ & $0.000$ & $0.748$ \\
\addlinespace[2pt]
\multicolumn{5}{l}{\emph{deletions, $k$ items}}\\
\quad $k{=}1$ & $0.842$ & $0.842$ & $0.000$ & $0.000$ \\
\quad $k{=}2$ & $0.684$ & $0.684$ & $0.000$ & $0.000$ \\
\quad $k{=}3$ & $0.575$ & $0.575$ & $0.000$ & $0.000$ \\
\addlinespace[2pt]
\multicolumn{5}{l}{\emph{insertions, $k$ duplicate items}}\\
\quad $k{=}1$ & $0.867$ & $0.867$ & $0.000$ & $0.000$ \\
\quad $k{=}2$ & $0.769$ & $0.769$ & $0.000$ & $0.000$ \\
\quad $k{=}3$ & $0.692$ & $0.692$ & $0.000$ & $0.000$ \\
\bottomrule
\end{tabular}

\end{table}

\subsection{Extrinsic study: setup}
\label{sec:results-setup}

We study the \OA{} in the role it was originally designed for: a deterministic reward (and,
optionally, a deterministic feedback signal) inside a prompt-optimization loop.

\paragraph{Optimizer and models.}
We use a single optimizer, GEPA~\cite{gepa}: our grid of datasets,
ablations, arms, and seeds makes a second optimizer computationally prohibitive,
and GEPA is a strong representative---state of the art (surpassing
MIPROv2~\cite{miprov2}), sample-efficient (up to $35\times$ fewer rollouts than
GRPO), and a standard optimizer of the popular DSPy framework~\cite{dspy}.
GEPA is a reflective prompt optimizer that
iteratively rewrites a task prompt. It maintains a Pareto frontier of candidate
prompts over the individual training instances and proposes each mutation from an
LLM reflection step that reads execution traces. 
Candidates are screened on small
reflection minibatches (size three) rather than full validation sweeps. 
The reflection (proposer) LM is a \emph{frozen} \mbox{GPT-5} model
(\texttt{gpt-5-2025-08-07})~\cite{gpt5}: pairing the reflector with
a strong frontier model is the recommended practice for reflective optimization and
the regime in which GEPA was developed~\cite{gepa}. The prompt being optimized drives
a separate, locally served \emph{task} model that emits the
structured JSON. We report two task models~\cite{gemma4} that bracket a wide capacity
range---the large, stronger \mbox{Gemma-4-26B-A4B-it} and the much smaller,
consumer-grade \mbox{Gemma-4-E4B-it}---to check that the qualitative findings are not
an artifact of one model's capacity.\footnote{As an additional robustness
check, we ran \mbox{Qwen3.5-35B-A3B} and \mbox{Qwen3.5-9B}~\cite{qwen35} as task
models on a subset of datasets. The qualitative conclusions were unchanged.}

\paragraph{Protocol.}
The datasets are described in Section~\ref{sec:datasets}. Each is split
into three disjoint partitions: a \emph{feedback} set, from which GEPA draws the
size-$3$ reflection minibatches; a \emph{Pareto} set, on which candidate prompts
are scored to maintain the frontier; and a held-out \emph{test} set, touched only
for the final evaluation.
For each
(dataset, model, arm, ablation) cell, we run GEPA under a fixed rollout budget of
$800$ metric calls for the synthetic Org2Graph and Facts2Order tasks and $600$ calls for
the real datasets, repeating over $10$ random seeds.
We aggregate the mean\,$\pm$\,sample standard deviation over the seeds.
Every experiment is seeded from a single minimal system prompt that specifies
only the required JSON output format and deliberately omits any hints for the
input-to-prediction transform; GEPA must discover that mapping purely from
\OA{}'s scalar score and, in the feedback arm, its textual feedback, isolating
whether the metric supplies enough optimization gradient. Verbatim seed
prompts for all datasets are listed in Appendix~\ref{app:prompts}.
The task model decodes under JSON-Schema--constrained decoding with an $8192$-token output cap,
greedily on Org2Graph (temperature~$0$) and at a
temperature of~$0.3$ on every other dataset; a run that degenerates into repeated tokens and fails to complete is retried up to twice
with the temperature raised by $0.2$ per attempt, and output that still fails to complete
counts as a failure (scored $0$; see ``Evaluation'' below).

\paragraph{Score and feedback arms.}
GEPA consumes whatever the metric returns. We compare two \emph{arms}.
In both arms, the scalar \OA{} score is GEPA's fitness signal, which drives
Pareto-frontier maintenance and candidate selection; the arms differ only in
what the reflection step additionally reads. In the
\textbf{score} arm, the metric returns only \OA{}'s scalar graded score; therefore, the
reflection model sees nothing but a number. In the \textbf{feedback} arm, it also
returns \OA{}'s deterministic, decomposable feedback string
(Section~\ref{sec:oa-feedback})---a ranked list of the $K=5$ most important repair operations with their
exact score deltas---in the natural-language reflection slot. The two arms share
everything else, so their difference isolates the value of the feedback.

\paragraph{Ablation axes.}
Each dataset isolates one of the two \OA{} extensions, and we treat that
extension as a control on GEPA's \emph{fitness function}. The question is not whether
the \OA{} \emph{can} measure a property (which was already established), but whether giving GEPA a reward that
\emph{perceives} that property yields a better optimized prompt than a reward that
is blind to it.
On the graph-extraction datasets, the axis is \textbf{referential alignment}
(Section~\ref{sec:oa-ref}): the \textbf{RA} schema marks the identifier and reference
fields, so the score is invariant to identifier relabeling---it judges a
graph up to isomorphism rather than penalizing a correct graph that merely names
its nodes differently---while the \textbf{plain} schema drops that machinery and
compares identifiers by value, so it cannot see relabeling at all. On the
ordered-output datasets, the axis is the sequence regime
(Section~\ref{sec:oa-lists}): the \textbf{sequence} schema perceives the output
order, while \textbf{set}, the order-agnostic Hungarian matching, rewards
membership only, and is blind to order. On each axis, one arm's reward sees the
target property and the other does not.

\paragraph{Evaluation.}
We always evaluate on held-out data with the \emph{property-sensitive}
schema (the relabel-invariant RA schema for the referential axis, the
\emph{sequence} schema for the order axis). Only these can detect the property in
question (graph isomorphism and correct order), and they are the richest scale common
to every dataset, including the synthetic graph and ordering tasks that have no
native metric. Holding this schema fixed, our quantity of interest is the
seed-paired contrast \textbf{RA\,$-$\,plain} and \textbf{sequence\,$-$\,set}: since
only the reward changes between the two conditions, the contrast attributes any
gain to the reward.

We understand that this estimate may not be fully honest, because the evaluation schema coincides with the
property-aware arm's own reward (RA, sequence), which can favor that arm. 
Therefore, we cross-check every contrast against a fully schema-independent, task-native
metric wherever one exists (defined per dataset in Appendix~\ref{app:native}); on
every real-world dataset that carries one, the native metric agrees in the direction
of the OA-score conclusion.

\paragraph{Statistical reporting.}
Every extrinsic contrast is seed-paired (the same $10$ seeds in both
conditions). We summarize its uncertainty with a $95\%$
bias-corrected-and-accelerated (BCa) bootstrap confidence interval over the
per-seed differences ($20{,}000$ resamples), reported inline for the
small-effect cells where the direction is not otherwise evident. We treat these
intervals descriptively---as effect-size estimates rather than a battery of
hypothesis tests---and throughout call a contrast \emph{detectable} when its
interval excludes zero and \emph{within noise} when it does not.

\paragraph{Research questions.}
We organize the results around three questions, and at this stage, we only
\emph{enumerate} the cases in which each extension helps; the dataset properties
that drive these patterns are discussed in Section~\ref{sec:discussion}.
\begin{itemize}
  \item \textbf{RQ1.} When does referential alignment (RA) improve the optimized
    prompt?
  \item \textbf{RQ2.} When does the order-sensitive (sequence) regime help over
    the order-agnostic (set) one?
  \item \textbf{RQ3.} When does the deterministic feedback help over the scalar
    score alone?
\end{itemize}

\begin{figure*}[t]
\centering
\includegraphics[width=\textwidth]{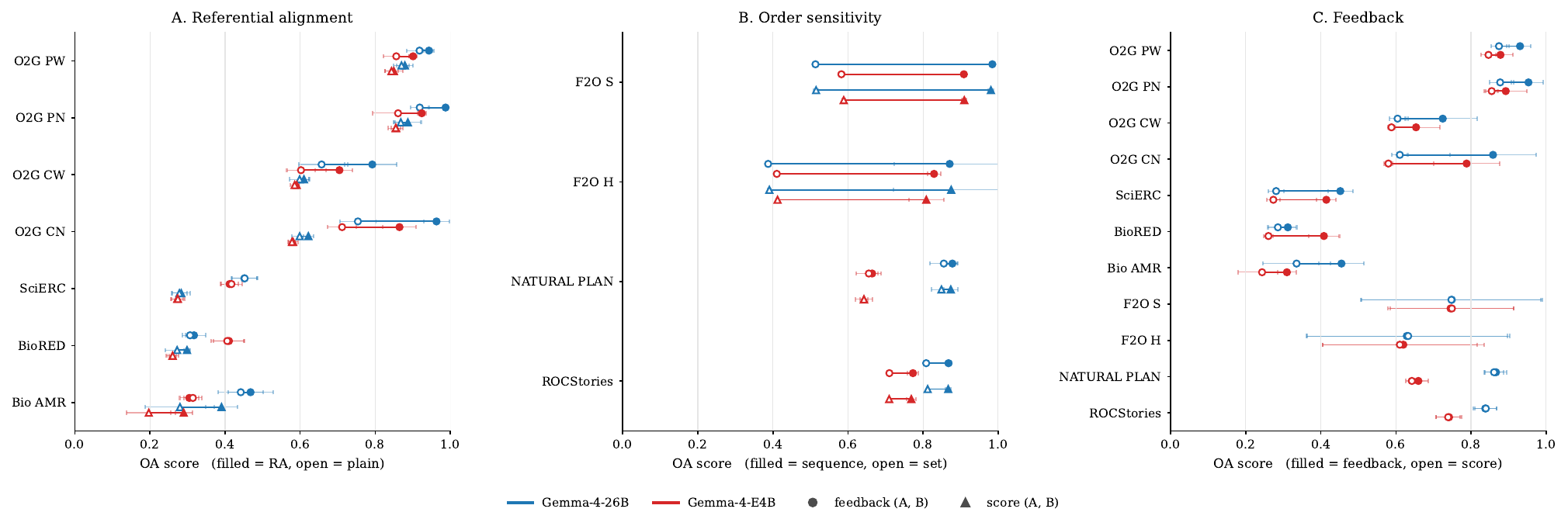}
\caption{The three contrast effects drawn as \emph{dumbbells} on the
absolute \OA{} score. Each series joins its two contrast
conditions (filled and open) with a connector whose
length shows the seed-paired effect. The
thin caps depict $\pm1$ std over seeds. 
Colour denotes the task model; in
panels~\textbf{(A,B)} marker shape denotes the reflection arm
(circle\,$=$\,feedback, triangle\,$=$\,score). \textbf{(A)}~RA vs.\ plain on the
\OA{} score under the RA schema. \textbf{(B)}~sequence vs.\ set on the sequence \OA{}
schema. \textbf{(C)}~feedback vs.\
score on each dataset's \OA{} evaluation schema, pooled over both ablations and shown
for every dataset on both axes. Dataset
codes: O2G\,$=$\,Org2Graph (P/C\,$=$\,plain/coded values, W/N\,$=$\,wide/narrow
vocabulary), F2O\,$=$\,Facts2Order (S/H\,$=$\,stated/hidden sort key).}
\label{fig:results-forest}
\end{figure*}

\subsection{Referential alignment}
\label{sec:results-ra}

Table~\ref{tab:results-ra} and Fig.~\ref{fig:results-forest}(A) compare the
RA and plain rewards across seven datasets: four synthetic Org2Graph
variants and three real-world graph-extraction tasks---SciERC, BioRED, and
Bio~AMR. The effect is sharply dataset-dependent. RA pays off on the synthetic
Org2Graph family, and there the gain grows with the difficulty of recovering the
graph: from $+0.03$--$0.07$ (PW to PN, Gemma-4-26B) on the \emph{plain} variants (readable
categorical values) to $+0.10$--$0.21$ (CW on Gemma-4-E4B to CN on Gemma-4-26B) on the \emph{coded} variants,
where the gold values are opaque codes the optimizer must learn a legend for.
Crucially, this advantage appears almost entirely in the feedback arm---under the
score arm the RA-vs-plain gap collapses to within noise
(Fig.~\ref{fig:results-forest}(A), open markers), so RA and feedback are not
independent levers but interact: the relabel-invariant reward only becomes
actionable once the feedback localizes which references are wrong. Both task
models exhibit the same pattern.

On SciERC and BioRED, the effect is far less pronounced, bordering on
noise: the seed-paired
RA\,$-$\,plain gap stays within noise in every cell but one
(Gemma-4-26B feedback/score $-0.001$/$+0.005$ on SciERC and $+0.011$/$+0.026$\footnote{The lone exception is the BioRED
Gemma-4-26B score arm: $+0.026$ (CI $[+0.012,+0.048]$), a small gain at the edge
of detectability.} on
BioRED). Although BioRED's gold entity ids are opaque concept accessions not
derivable from the text (which in principle should force referential
routing), its entities stay well individuated by their surface mentions and its
gold relations are sparse, therefore, exactly as in SciERC, the relabel-invariant reward
has little routing left to recover. Bio AMR provides a different picture: here, the
RA advantage approaches the coded Org2Graph variants in
magnitude. Its dense AMR graphs reuse the same concept
symbol across many nodes, so node content individuates almost nothing, and the
score turns entirely on reference routing. Here, though, the pattern
flips: the RA advantage lives in the \emph{score} arm and all but disappears under
feedback (Gemma-4-26B feedback/score $+0.026$/$+0.111$; $-0.010$/$+0.093$ on
Gemma-4-E4B)---the mirror image of coded Org2Graph. The two arms diverge over
\emph{where} the routing signal can travel. A plain score is blind to routing on
AMR: edges are keyed on gold node identifiers the model never sees, so the scalar
rewards only node content and leaves routing untouched---and RA reward supplies
exactly that missing edge credit, hence its large score-arm gain. Under feedback, 
the gap closes instead, because OA's feedback \emph{text} still spells out the
mis-routed edges even when the plain score ignores them; the reflection model
uses that referential information in the prompt even though GEPA's plain-score
selection never rewards it, so plain and RA feedback converge on the same graph.
In effect, plain feedback has already done the RA's job. 

The contrast with
coded Org2Graph turns on a single difference: AMR node content is genuine,
text-derivable semantics (e.g., \texttt{express-03}, \texttt{protein},
\texttt{cell}), whereas Org2Graph's coded identifiers are arbitrary
codes with no content handle. That semantic content is what lets plain feedback leak
the routing here: the fixes it lists are concrete content corrections the
model can actually act on (e.g., ``add a node for this entity''), and
carrying them out repairs the graph structure as a by-product---so under feedback
plain already teaches routing and the RA advantage collapses there. On coded Org2Graph
plain feedback can only report unfollowable code mismatches, so relabeling is
recoverable only through RA feedback and the advantage instead survives there,
shifting the collapse to the score arm.

\begin{table*}[t]
\centering
\caption{Referential-alignment axis. Each ablation optimizes the prompt
(via GEPA) against its own \OA{} reward as fitness (using RA or plain schema, 
Section~\ref{sec:results-setup})
All configurations are then evaluated using the RA schema, 
measured on each dataset's held-out test
split (mean\,$\pm$\,seed std, $10$ seeds); the larger of RA/plain
within each arm is \textbf{bold}. The final pair is the feedback\,$-$\,score
improvement per ablation (means only). O2G\,$=$\,Org2Graph variants
(P/C\,$=$\,plain/coded values, W/N\,$=$\,wide/narrow vocabulary).}
\label{tab:results-ra}
\footnotesize
\setlength{\tabcolsep}{4pt}
\begin{tabular}{ll cc cc cc}
\toprule
 & & \multicolumn{2}{c}{feedback} & \multicolumn{2}{c}{score} & \multicolumn{2}{c}{feedback $-$ score} \\
\cmidrule(lr){3-4}\cmidrule(lr){5-6}\cmidrule(lr){7-8}
Dataset & Model & RA & plain & RA & plain & RA & plain \\
\midrule
O2G PW & Gemma-4-26B & \textbf{0.943\,$\pm$\,0.014} & 0.918\,$\pm$\,0.034 & \textbf{0.879\,$\pm$\,0.021} & 0.870\,$\pm$\,0.020 & 0.064 & 0.048 \\
 & Gemma-4-E4B & \textbf{0.901\,$\pm$\,0.011} & 0.856\,$\pm$\,0.034 & \textbf{0.850\,$\pm$\,0.024} & 0.843\,$\pm$\,0.017 & 0.051 & 0.013 \\
\midrule
O2G PN & Gemma-4-26B & \textbf{0.987\,$\pm$\,0.003} & 0.919\,$\pm$\,0.024 & \textbf{0.887\,$\pm$\,0.035} & 0.868\,$\pm$\,0.019 & 0.100 & 0.050 \\
 & Gemma-4-E4B & \textbf{0.924\,$\pm$\,0.012} & 0.861\,$\pm$\,0.067 & \textbf{0.855\,$\pm$\,0.013} & 0.855\,$\pm$\,0.020 & 0.069 & 0.006 \\
\midrule
O2G CW & Gemma-4-26B & \textbf{0.792\,$\pm$\,0.065} & 0.657\,$\pm$\,0.061 & \textbf{0.610\,$\pm$\,0.013} & 0.599\,$\pm$\,0.026 & 0.182 & 0.059 \\
 & Gemma-4-E4B & \textbf{0.705\,$\pm$\,0.035} & 0.602\,$\pm$\,0.038 & \textbf{0.591\,$\pm$\,0.008} & 0.585\,$\pm$\,0.011 & 0.114 & 0.017 \\
\midrule
O2G CN & Gemma-4-26B & \textbf{0.963\,$\pm$\,0.034} & 0.754\,$\pm$\,0.048 & \textbf{0.622\,$\pm$\,0.013} & 0.599\,$\pm$\,0.022 & 0.341 & 0.155 \\
 & Gemma-4-E4B & \textbf{0.865\,$\pm$\,0.044} & 0.712\,$\pm$\,0.038 & \textbf{0.581\,$\pm$\,0.013} & 0.578\,$\pm$\,0.009 & 0.283 & 0.133 \\
\midrule
SciERC & Gemma-4-26B & 0.451\,$\pm$\,0.034 & \textbf{0.452\,$\pm$\,0.034} & \textbf{0.283\,$\pm$\,0.023} & 0.278\,$\pm$\,0.020 & 0.168 & 0.174 \\
 & Gemma-4-E4B & 0.412\,$\pm$\,0.024 & \textbf{0.417\,$\pm$\,0.028} & 0.273\,$\pm$\,0.016 & \textbf{0.274\,$\pm$\,0.019} & 0.139 & 0.143 \\
\midrule
BioRED & Gemma-4-26B & \textbf{0.317\,$\pm$\,0.031} & 0.307\,$\pm$\,0.011 & \textbf{0.299\,$\pm$\,0.008} & 0.272\,$\pm$\,0.032 & 0.019 & 0.034 \\
 & Gemma-4-E4B & \textbf{0.411\,$\pm$\,0.041} & 0.406\,$\pm$\,0.044 & 0.260\,$\pm$\,0.010 & \textbf{0.260\,$\pm$\,0.016} & 0.150 & 0.145 \\
\midrule
Bio AMR & Gemma-4-26B & \textbf{0.468\,$\pm$\,0.060} & 0.442\,$\pm$\,0.060 & \textbf{0.391\,$\pm$\,0.043} & 0.280\,$\pm$\,0.092 & 0.077 & 0.162 \\
 & Gemma-4-E4B & 0.305\,$\pm$\,0.026 & \textbf{0.314\,$\pm$\,0.025} & \textbf{0.290\,$\pm$\,0.023} & 0.197\,$\pm$\,0.059 & 0.015 & 0.117 \\
\bottomrule
\end{tabular}

\end{table*}

The Org2Graph sweep over four fixed variants spanning
obfuscation $\times$ width (\emph{plain}/\emph{coded} $\times$
\emph{wide}/\emph{narrow}; Section~\ref{sec:dataset-org2graph}) makes the
difficulty scaling explicit; the per-variant feedback-arm numbers are the four
O2G rows of Table~\ref{tab:results-ra} and the top four series in
Fig.~\ref{fig:results-forest}(A). The family varies on two axes: identifier
obfuscation (plain readable values vs.\ opaque codes) and vocabulary size (wide
vs.\ narrow code pools), with graph topology held fixed along the obfuscation
axis so that plain and coded variants differ only in surface form. Obfuscation
is the dominant driver: switching from plain to coded values widens the RA
advantage several-fold on both models (RA\,$-$\,plain rising from $+0.03$/$+0.07$
(PW/PN) on the plain variants to $+0.14$/$+0.21$ (CW/CN) on the coded variants for Gemma-4-26B,
and from $+0.05$/$+0.06$ to $+0.10$/$+0.15$ for Gemma-4-E4B), because once the
surface tokens carry no signal the only way to credit a correct edge is through
the inferred identifier bijection. Narrowing the vocabulary, which forces more
property-identical records and thus leans harder on the structural tie-break,
adds a smaller further increment. The advantage is monotone in our informal
difficulty ranking.

\subsection{Order sensitivity}
\label{sec:results-order}

Table~\ref{tab:results-order} and Fig.~\ref{fig:results-forest}(B) compare the
sequence and set rewards, both scored on the sequence \OA{} schema. Order
sensitivity helps precisely when the position of each element is itself part of the
answer. Evaluated on the sequence schema, optimizing with
the sequence reward drives the score from approximately $0.83$ to
$0.98$, against $0.39$ to
$0.58$ for the set reward---a sequence$-$set gap between
$+0.33$ (F2O~S, Gemma-4-E4B) and $+0.48$ (F2O~H, Gemma-4-26B) on every variant
and both models. The effect
carries over to our real ordering tasks to a lesser extent: on ROCStories, the sequence reward
increases the sequence-schema score by approximately $0.06$. On NATURAL PLAN the gain is
marginal---detectable but practically negligible: the
sequence$-$set gap stays $\le+0.025$ across all four cells---about $+0.02$--$0.025$ on
Gemma-4-26B (both arms) and $\le+0.009$ on Gemma-4-E4B.\footnote{Although
small, the Gemma-4-26B gap is statistically detectable: $+0.025$ ($95\%$ CI
$[+0.008,+0.045]$) under the score reward and $+0.023$ ($95\%$ CI $[+0.005,+0.046]$)
under feedback, both excluding zero; on Gemma-4-E4B it is indistinguishable from zero.}

The contrast with Facts2Order explains this result. The NATURAL PLAN itinerary is
also strictly ordered (Section~\ref{sec:dataset-natural-plan}), solving the
task already entails finding the order: to satisfy the direct-flight graph and the
day budget, the model must determine which city follows which, and once it has, it
emits the cities in that sequence as the natural way to present the answer---even
though the order-agnostic set reward never asks it to. Therefore, ordering is a
by-product of the solution, not a separate thing to optimize, so rewarding it adds
little. Specifically, the set-optimized prompt already scores $0.855$/$0.655$ on the
sequence schema, only marginally below the sequence-optimized $0.878$/$0.664$. Facts2Order
is the mirror image: the items are handed to the model already solved, and only their
arrangement by the sort key is in question, so the set reward leaves ordering
unsolved (on F2O~H the set-optimized prompt collapses to $0.387$ on the sequence
schema against $0.871$ for sequence-optimized), and ordering is the entire task.

\begin{table*}[t]
\centering
\caption{Order-sensitivity axis. Each ablation optimizes the prompt (via
GEPA) against its own \OA{} reward as fitness (using sequence or set schema,
Section~\ref{sec:results-setup}), and all configurations are then evaluated using the
sequence schema, measured on each dataset's held-out test split
(mean\,$\pm$\,seed std, $10$ seeds); the larger of sequence/set within each arm is
\textbf{bold}. The final pair is the feedback\,$-$\,score improvement per ablation
(means only). F2O\,$=$\,Facts2Order
variants (S/H\,$=$\,stated/hidden sort key).}
\label{tab:results-order}
\footnotesize
\setlength{\tabcolsep}{4pt}
\begin{tabular}{ll cc cc cc}
\toprule
 & & \multicolumn{2}{c}{feedback} & \multicolumn{2}{c}{score} & \multicolumn{2}{c}{feedback $-$ score} \\
\cmidrule(lr){3-4}\cmidrule(lr){5-6}\cmidrule(lr){7-8}
Dataset & Model & sequence & set & sequence & set & sequence & set \\
\midrule
F2O S & Gemma-4-26B & \textbf{0.984\,$\pm$\,0.005} & 0.513\,$\pm$\,0.004 & \textbf{0.980\,$\pm$\,0.007} & 0.515\,$\pm$\,0.006 & 0.004 & -0.002 \\
 & Gemma-4-E4B & \textbf{0.909\,$\pm$\,0.007} & 0.582\,$\pm$\,0.008 & \textbf{0.910\,$\pm$\,0.006} & 0.589\,$\pm$\,0.005 & -0.001 & -0.006 \\
\midrule
F2O H & Gemma-4-26B & \textbf{0.871\,$\pm$\,0.148} & 0.387\,$\pm$\,0.003 & \textbf{0.875\,$\pm$\,0.153} & 0.390\,$\pm$\,0.004 & -0.004 & -0.003 \\
 & Gemma-4-E4B & \textbf{0.829\,$\pm$\,0.017} & 0.410\,$\pm$\,0.005 & \textbf{0.808\,$\pm$\,0.046} & 0.412\,$\pm$\,0.003 & 0.021 & -0.002 \\
\midrule
NATURAL PLAN & Gemma-4-26B & \textbf{0.878\,$\pm$\,0.014} & 0.855\,$\pm$\,0.036 & \textbf{0.874\,$\pm$\,0.019} & 0.849\,$\pm$\,0.026 & 0.004 & 0.006 \\
 & Gemma-4-E4B & \textbf{0.664\,$\pm$\,0.016} & 0.655\,$\pm$\,0.033 & 0.642\,$\pm$\,0.022 & \textbf{0.642\,$\pm$\,0.010} & 0.022 & 0.013 \\
\midrule
ROCStories & Gemma-4-26B & \textbf{0.868\,$\pm$\,0.008} & 0.808\,$\pm$\,0.004 & \textbf{0.867\,$\pm$\,0.008} & 0.812\,$\pm$\,0.002 & 0.001 & -0.004 \\
 & Gemma-4-E4B & \textbf{0.773\,$\pm$\,0.015} & 0.710\,$\pm$\,0.004 & \textbf{0.769\,$\pm$\,0.013} & 0.709\,$\pm$\,0.002 & 0.004 & 0.001 \\
\bottomrule
\end{tabular}

\end{table*}

The two Facts2Order variants contrast the sort-key visibility
(stated\,$\to$\,hidden; Section~\ref{sec:dataset-facts2order}). 
Both draw the item count from $N\in\{4,5,7,9\}$, the key type from \{integer, date, ordinal\}, and
$0$ or $2$ irrelevant distractor sentences. In the stated variant~(S), each item
carries a single key clause; in the hidden variant~(H), each item additionally
carries $2$--$3$ numeric decoy clauses, shuffled per item so the sort field must
be discovered from \OA{} feedback (Fig.~\ref{fig:facts2order-sample},
Appendix~\ref{app:intrinsic}, shows a worked hidden-key instance---the most
complex configuration). Their per-variant numbers
are the two F2O rows in Table~\ref{tab:results-order} and the corresponding series
in Fig.~\ref{fig:results-forest}(B). Unlike the
RA sweep, where the advantage only emerges as extraction grows harder, the
sequence$-$set gap is large on \emph{every} variant ($+0.33$--$0.48$): when the answer
is itself an ordering, the order-agnostic set reward leaves the bulk of the signal on
the table regardless of difficulty. Hiding the sort key instead moves the
\emph{absolute} score and the stability of optimization: it lowers the sequence
score and, on Gemma-4-26B, leaves it seed-unstable ($0.871\pm0.148$) under both
rewards alike---a score-vs-feedback question taken up in
Section~\ref{sec:results-feedback}, not the sequence-vs-set effect.

\subsection{Feedback}
\label{sec:results-feedback}

Across all 11 dataset\,$\times$\,axis cells, the
feedback\,$-$\,score columns of Tables~\ref{tab:results-ra}
and~\ref{tab:results-order} and the summary in
Fig.~\ref{fig:results-forest}(C) indicate that feedback helps in proportion
to how much structure the task hides from a scalar reward. It is decisive where
the optimizer must \emph{discover} something a number cannot convey: a 
value-to-code legend, a relation schema, or a strict output format. SciERC is
the starkest case: the score arm collapses to within noise, whereas feedback
recovers a working extractor. The same mechanism drives the large gains in the
coded Org2Graph variants and Bio~AMR. Where the format is simple and a scalar
gradient already locates the error, the feedback adds nothing. 
The order-sensitivity axis is the cleanest negative. 
Across all four
ordered-output datasets, the feedback\,$-$\,score gap remains within noise on both
arms and both models (Table~\ref{tab:results-order}); the two largest movements
are a faint $+0.022$ on NATURAL PLAN and $+0.021$ on F2O~H, both on the smaller
Gemma-4-E4B under the sequence reward.

\begin{figure*}[t]
\centering
\includegraphics[width=\textwidth]{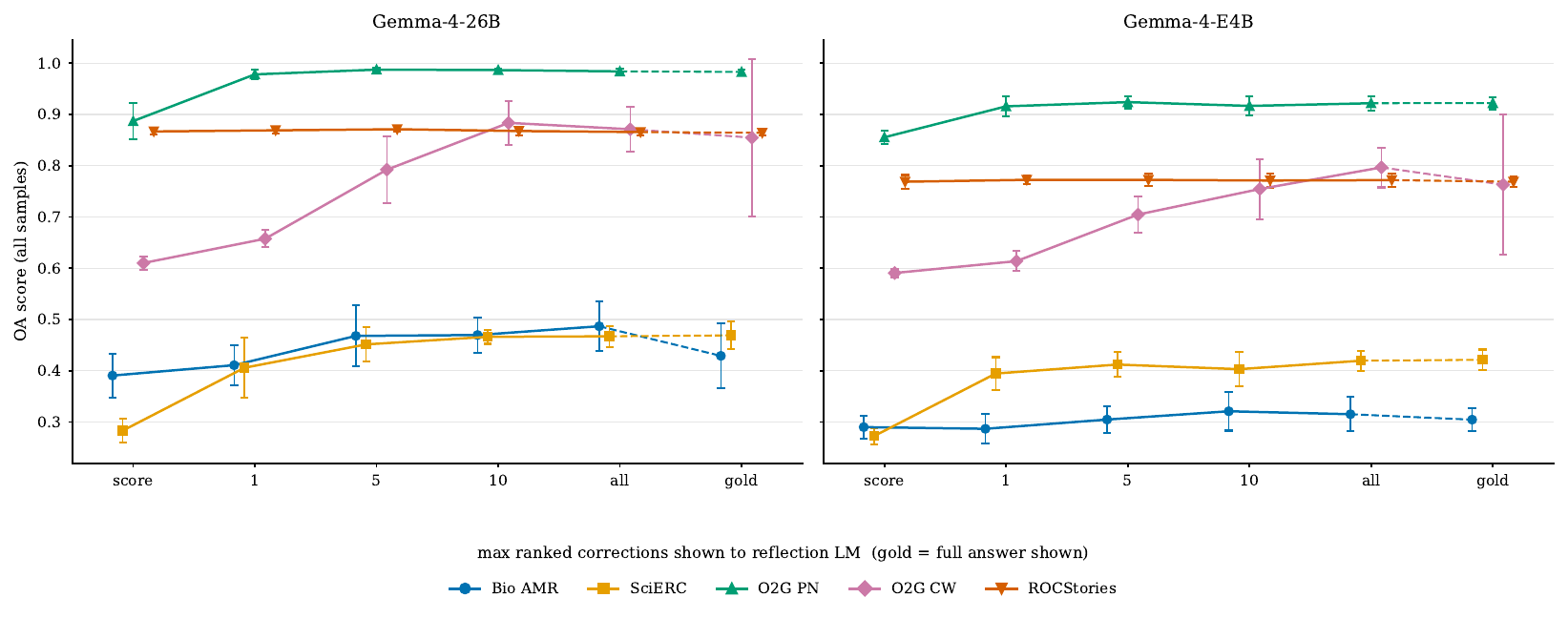}
\caption{Feedback-breadth sweep. The scalar reward is held
\emph{identical} across all arms; only the amount of ranked-correction
\emph{items} the OA metric shows the GEPA reflection LM varies with the
correction cap $K$ (Section~\ref{sec:oa-feedback}), from none
(\texttt{score}, $K{=}0$) through $K\,{=}\,1$, $5$ (the OA default), and $10$,
to all uncapped corrections ($K{=}\infty$). 
The \texttt{gold} column is an additional baseline that appends the full serialized gold answer. 
Each point is the mean GEPA-optimized holdout OA score over seeds
(whiskers $\pm1$ std); the per-dataset $x$ positions are slightly offset to reduce
overlap. Both Gemma models are shown.}
\label{fig:feedback-sweep}
\end{figure*}

Fig.~\ref{fig:feedback-sweep} resolves the contrast into a breadth
curve: holding the scalar reward identical, it varies only the number of ranked
corrections $K$ the \OA{} metric shows the reflection LM. We run the full sweep
on the five cells that most sharply discriminate the arms---one plain and one
coded Org2Graph variant rather than all four, the two real graph tasks where
the structure is most hidden (SciERC, Bio~AMR), and ROCStories as the order-axis
control. The shape of the curve tracks difficulty: on SciERC and the plain
O2G~PN, it is steeply concave, a single ranked correction ($K{=}1$) already
captures most of the gain, whereas on the harder coded O2G~CW and the dense
Bio~AMR, it climbs gradually and most of the gain arrives only by $K{=}5$--$10$. 
On ROCStories it remains flat throughout, mirroring the null result above.

The \texttt{gold} column appends the full serialized gold answer in the
reflection slot---an upper-information control rather than a usable reward
signal. It is not a standard input: handed the answer, the reflection LM must
itself work out which parts of its output were wrong, in effect performing
LLM-as-judge, and on a more verbose prompt than any deterministic reward
would emit. It is also markedly less stable across the seeds on the hardest
cells: on the coded O2G~CW the \texttt{gold} arm spans $\pm0.15$, against
$\pm0.05$ for the capped feedback arms. Its purpose is to bound the value of feedback information from above,
and \OA{} approaches that bound with little feedback.
The default $K{=}5$ arm already nears \texttt{gold} on most cells, and the uncapped arm meets or exceeds it on the
harder ones (Bio~AMR, the coded Org2Graph variants, SciERC). 
The fact that a short ranked list rivals showing the entire answer is direct evidence the gains come from
\OA{}'s \emph{prioritized, localized} corrections, not from answer leakage. It
also shows that $K{=}5$, the cap used throughout our main experiments, was
conservative rather than optimal, whereas on the hardest cells, the curve is still
climbing at $K{=}5$, so a larger cap would have scored higher at the cost of a
longer feedback---a trade we did not tune per dataset.

\section{Discussion}
\label{sec:discussion}

The intrinsic study
(Section~\ref{sec:intrinsic}) shows that each new mechanism is the \emph{uniquely
correct} measurement of its target property: referential alignment (RA)
separates damaged graphs from clean ones almost perfectly, where
the reference-unaware \emph{plain} ablation, blind to reference rerouting, does
not, and the \emph{sequence} regime is the only scorer that grades
order at all. The extrinsic study asks whether a reward that
\emph{perceives} that property yields a better prompt, and there the answer is
conditional. The lesson is that \textbf{a faithful measurement is not
automatically a useful reward}: a property-aware reward helps the search only when the
property is (i)~\emph{not already recovered as a by-product} of solving the task, 
and (ii)~\emph{actionable} by the optimizer. The first condition governs when RA
(RQ1) and the sequence regime (RQ2) help, and the second condition connects both to
deterministic feedback (RQ3).

\textbf{RQ1.} RA helps precisely when node content fails to individuate
records, so the score must turn on reference routing rather than on attributes:
coded Org2Graph variants, whose identifiers are opaque codes, and Bio~AMR,
whose dense graphs reuse the same concept symbol across many nodes. Where surface
mentions already individuate entities and relations are sparse (SciERC, BioRED),
routing is solved before RA is consulted, and the gap becomes small. 
In the synthetic family, the advantage is monotone in extraction
difficulty, widening as readable values become opaque codes, with
obfuscation the dominant driver and vocabulary size the smaller one. This is not
``RA fails on real data'': RA is the only setting
under which a rerouted edge that leaves attributes intact is visible to the score
at all, and it pays off extrinsically on Bio~AMR, where the same gain
carries over in sign to AMR's own Smatch metric (Appendix~\ref{app:native}). What we could not find is a naturally
occurring corpus reproducing Org2Graph's adversarial regime---many property-twin
records behind obfuscated ids---under which RA's payoff is largest; whether such
``realistic'' data exists in the wild is an open question that normalized
relational databases, whose surrogate keys carry no meaning, may answer
(Section~\ref{sec:future-work}).
Several of the datasets that we spot-checked behaved consistently. 
RA and feedback are also not independent levers: on coded
Org2Graph, the RA advantage lives in the feedback arm and vanishes under the
scalar score, whereas on Bio~AMR, it is the mirror image. The difference is that
AMR concepts are genuine, text-derivable semantics, so even plain feedback lists
content fixes that repair structure as a by-product, whereas coded identifiers
carry no such handle and are recoverable only through RA feedback.

\textbf{RQ2.} The sequence reward is helpful only when the arrangement
\emph{is} the deliverable. On Facts2Order, where items arrive already solved and
only their ordering is in question, the sequence reward beats the order-agnostic
set reward by a wide margin for every variant. NATURAL PLAN is the contrast: satisfying the
flight graph and day budget already forces the city order, which the model then
emits as the natural presentation even though the set reward never asks for it,
so the sequence reward adds only a marginal gain; ROCStories sits between with a modest
gain. Hiding the sort key lowers the \emph{absolute} score and
its seed stability under \emph{both} rewards alike: a difficulty effect, not a
sequence-vs-set one. As with RA, this neutrality is a property of the
data, not the extension failing: where correct ordering is inherent to the task,
the sequence reward has little extra to teach, yet it remains the only scorer that
detects an incorrect order when one occurs. Order sensitivity thus helps as a reward
when the output order is itself part of the answer, and is neutral when a correct
solution fixes it for free.

\textbf{RQ3.} Deterministic feedback is the most consistently valuable
mechanism, and its value scales with how much structure the task hides from a
scalar reward. It is decisive where the optimizer must \emph{discover} something
a number cannot convey---a value-to-code legend, a relation schema, a strict
output format: on SciERC the score arm collapses to within noise while feedback
recovers a working extractor, and the same drives the gains on coded Org2Graph
and Bio~AMR. The feedback-breadth sweep
(Fig.~\ref{fig:feedback-sweep}) rules out the obvious confound: \OA{}'s short
ranked list of localized repairs \emph{matches} the \texttt{gold} baseline
(which appends the entire answer) on most cells and \emph{exceeds} it on the
hardest ones, while appending less text, direct evidence the gains come from
prioritized, localized corrections rather than answer leakage.

The two checks guard against artifacts. Task-model capacity: the two Gemma
models bracket a wide range, and capacity shifts magnitudes and absolute scores
but not the \emph{sign or ordering} of any contrast whose $95\%$ CI excludes zero.
Because the reward GEPA optimized was always an \OA{} schema, we cross-check
against a schema-independent native metric on every real-world dataset that
carries one (Appendix~\ref{app:native}): the native verdict agrees in direction
with the headline conclusions, diverging only in informative ways.

For practitioners, the advice is compact and shares one shape: each extension is
the correct measurement of its property, and as a reward, each only helped or stayed
neutral in our experiments---never a meaningful loss on any \OA{}-score
contrast\footnote{The one exception is negligible: on the ROCStories set arm,
feedback trails score by $-0.004$ ($95\%$ bootstrap CI $[-0.006, -0.000]$).}---so the practical question
is mainly where each \emph{pays off most}. Enable RA for any (hyper)graph-structured
output: it is correct whenever quality is judged up to isomorphism, degrades to
plain comparison in the worst case, and pays off most when identifiers are arbitrary
and content does not individuate records. Enable the sequence regime whenever the 
output order is part of correctness---there it never hurts and pays off most when
ordering is the deliverable. Finally deterministic feedback is preferred by default:
it is the broadest-paying lever, costs no extra model call, and pays off most where
the task hides structure from a scalar reward. Unlike an LLM judge, it is
reproducible and auditable across any number of rollouts (Section~\ref{sec:introduction}).

\section{Limitations}
\label{sec:limitations}
The main limitations fall into two categories: intrinsic properties
of the metric and the scope of our empirical evaluation.
\begin{itemize}
  \item \textbf{Ordinal, not calibrated.} The score is ordinal rather
    than calibrated: $0$ and $1$ are anchored and scores are comparable, but an
    intermediate value such as $0.51$ is not a literal fraction of
    correctness---a property \OA{} shares with many similarity metrics.
  \item \textbf{Approximate identifier bijection.} Referential alignment
    recovers the gold--candidate identifier bijection by Hungarian assignment on
    masked attributes, tie-broken by $1$-WL color refinement
    (Section~\ref{sec:oa-ref}), which is sound but incomplete approximation to graph
    isomorphism. On graphs, $1$-WL cannot separate, the inferred bijection---and
    thus the score---can be suboptimal; higher-order $k$-WL would tighten it
    (Section~\ref{sec:future-work}).
  \item \textbf{Self-referential and mutually-referential scopes.} Referential
    alignment resolves scopes in dependency order, so mutually referential
    scopes (or scope referencing itself) form a cycle with no valid order and
    fall back to property-only alignment, discarding the reference-value signal.
    Resolving such cases consistently is a fixed-point problem (akin to
    collective entity resolution~\cite{bhattacharya2007collective}), which we
    leave for future work. In practice, many such cycles can be sidestepped by
    \emph{reifying} the offending references as separate junction (link)
    table---pure reference-carrier records with no \texttt{idScope} of their
    own, the relational model's \emph{relationship relation}~\cite{chen1976er}---which
    restores an acyclic scope order, at the cost
    of treating each edge as an independently matchable record rather than a
    per-record attribute.
  \item \textbf{Scope of the empirical study.} We use a single optimizer
    (GEPA), a single reflection LM (a strong frontier model, which is
    best practice), and two task models chosen partly by our available
    infrastructure; whether the extrinsic findings transfer to other
    prompt-optimization frameworks or model families is
    untested, although our partial Qwen robustness check was consistent.
  \item \textbf{Limited hyperparameter exploration.} The cost of the
    full grid prevented us from sweeping alternative schema-construction
    strategies, GEPA hyperparameters, and \OA{} metaparameters (e.g., the
    feedback list length $K$).
  \item \textbf{Not a benchmark study.} Our aim was to demonstrate
    \OA{}'s general properties in a prompt-optimization setup, not to set
    state-of-the-art baselines on the datasets; as a highly configurable metric,
    \OA{} may need per-task schema tuning for the best performance.
\end{itemize}

\section{Conclusion}
\label{sec:conclusion}
We presented \OA{}, an open-source Python library that scores
structured LLM outputs against a gold reference deterministically by
recursively aligning their JSON trees and awarding partial credit at the
granularity a schema declares. Because it is configured entirely through JSON
Schema extensions, adapting the \OA{} to a new task is a matter of annotating a
schema rather than writing code, and it drops directly into existing
prompt-optimization frameworks such as DSPy, GEPA, and TextGrad as a
reproducible, auditable, and model-free alternative to an LLM judge. Its
central contribution, \emph{referential alignment}, makes the score
invariant to identifier relabeling,  so that cross-referenced records---graphs and
hypergraphs---can be compared up to isomorphism, while a complementary
order-sensitive \emph{sequence} regime targets ranking, and the
same alignment emits ranked, localized repair suggestions at no extra model
cost. Used as the reward inside GEPA across synthetic and real-world
datasets, \OA{} helped or stayed neutral, never a meaningful loss: referential alignment pays off
most when identifiers are arbitrary and content fails to individuate records, 
the sequence regime when the output order is itself the deliverable, and
deterministic feedback proved the broadest-paying lever. These findings
translate into compact, deployable guidance---enable referential alignment for
any (hyper)graph output, the sequence regime whenever order is part of
correctness, and prefer deterministic feedback by default---positioning \OA{} as
a practical, drop-in measurement and feedback signal for building and optimizing
the structured-output pipelines on which LLM applications increasingly depend.

\section{Future Work}
\label{sec:future-work}
Three directions follow most directly from our findings and the
limitations above:
\begin{itemize}
  \item \textbf{Referential alignment for relational data.} The normalized
    relational databases reproduce Org2Graph's adversarial regime
    (Section~\ref{sec:discussion}): surrogate keys that carry no
    content-derived meaning, with foreign
    keys that are exactly \OA{}'s \texttt{idScope}/\texttt{ref} structure, with
    junction tables as the purest property-twin case. This suggests the application of \OA{}
    to text-to-database extraction and database-state diffing, building on the
    record-linkage lineage (Section~\ref{sec:rel-metrics}), and collective entity
    resolution~\cite{bhattacharya2007collective}.
  \item \textbf{Differentiable surrogates.} \OA{}'s score is derived from a
    discrete Hungarian assignment; therefore, it can only act as a black-box scalar or
    textual reward (Section~\ref{sec:discussion}). Relaxing the assignment into a soft,
    entropy-regularized (Sinkhorn-style) matching may yield a differentiable
    surrogate, opening \OA{} to gradient-based optimization and end-to-end
    fine-tuning rather than search alone.
  \item \textbf{Stronger structural tie-breaking.} Replace 1-WL with
    higher-order $k$-dimensional Weisfeiler--Leman to resolve its known blind spots
    (e.g.,  one $6$-cycle vs.\ two disjoint $3$-cycles), trading the near-linear cost
    for the combinatorial-in-$k$ scaling of $k$-WL~\cite{morris2019kgnn}.
\end{itemize}

\appendices

\section{Schema Keyword Reference}
\label{app:keywords}

Table~\ref{tab:oa-keywords} maps the mathematical notation used in
Section~\ref{sec:object-aligner} to the concrete schema keywords accepted by the
implementation. The schema reuses the JSON~Schema surface syntax; standard
validation keywords (\texttt{required}, \texttt{enum}, \texttt{minItems},
\dots) are honored when validating the candidate but do not affect the score.
Custom primitive comparators are registered on the aligner and referenced by
name. Table~\ref{tab:oa-keywords} lists only the \OA{} scoring
extensions.

\begin{table}[h]
\caption{Notation of Section~\ref{sec:object-aligner} versus schema keywords.
Defaults in the last column.}
\label{tab:oa-keywords}
\centering
\footnotesize
\begin{tabular}{@{}l l l@{}}
\toprule
\textbf{Concept / symbol} & \textbf{Keyword} & \textbf{Default} \\
\midrule
\multicolumn{3}{@{}l}{\textit{Primitive leaf} (string / number / boolean)} \\
\quad comparator $\sleaf$ (string) & \texttt{score} & \texttt{jaro} \\
\quad comparator $\sleaf$ (number) & \texttt{score} & \texttt{invdiff} \\
\quad threshold $\tau$ & \texttt{threshold} & $0$ \\
\quad empty-value score (asymmetric) & \texttt{nullScore} & $0$ \\
\midrule
\multicolumn{3}{@{}l}{\textit{Map} (object)} \\
\quad key comparator $\kappa$ & \texttt{keyScore} & \texttt{jaro} \\
\quad key threshold $\tau_K$ & \texttt{keyThreshold} & $0$ \\
\quad key weight $w_K$ & \texttt{keyImportance} & $0$ \\
\quad value weight $w_V$ & \texttt{valueImportance} & $1$ \\
\quad per-property weight $\omega_i$ & \texttt{valueWeight} & $1$ \\
\midrule
\multicolumn{3}{@{}l}{\textit{Sequence} (array)} \\
\quad order regime & \texttt{order} & \texttt{fixed} \\
\quad\quad order-agnostic & \quad\texttt{"align"} & \\
\quad\quad order-sensitive & \quad\texttt{"fixed"} & \\
\quad prefix weights & \texttt{prefixWeights} & $1$ \\
\quad penalize missing ($n_{\mathrm{miss}}$) & \texttt{ignoreMissing} & penalize \\
\quad penalize excess ($n_{\mathrm{exc}}$) & \texttt{ignoreExcess} & penalize \\
\quad prefix importance $w_p$ & \texttt{prefixImportance} & --- \\
\quad tail importance $w_r$ & \texttt{restImportance} & --- \\
\midrule
\multicolumn{3}{@{}l}{\textit{Referential} (primitive inside a sequence)} \\
\quad identifier field & \texttt{idScope} & --- \\
\quad reference field & \texttt{ref} & --- \\
\bottomrule
\end{tabular}
\end{table}

The \OA{} scoring extensions that appear in the worked schema
examples throughout the paper are glossed below: the standard JSON~Schema
keywords carry their usual meaning. The authoritative specification is the
official documentation at
\url{https://github.com/aic-factcheck/object_aligner}.

\begin{itemize}
\item \texttt{score} --- primitive comparator (\texttt{exact},
  \texttt{jaro}, \texttt{invdiff}, \dots).
\item \texttt{threshold} --- minimum primitive similarity to count as
  a match; below it the pair scores $0$.
\item \texttt{order} --- array regime: \texttt{"align"}
  (order-agnostic Hungarian matching) or \texttt{"fixed"} (order-sensitive).
\item \texttt{keyImportance}\,/\,\texttt{valueImportance} ---
  relative weight of an object's keys vs.\ its values in the node score.
\item \texttt{valueWeight} --- per-property weight within an object.
\item \texttt{prefixWeights} --- per-position weights for the
  \texttt{prefixItems} entries.
\item \texttt{ignoreExcess} --- do not penalize extra, unmatched
  candidate elements.
\item \texttt{idScope} --- declares a primitive as an identifier
  within a named scope (enabling referential alignment).
\item \texttt{ref} --- declares a primitive as a reference resolved
  against a named \texttt{idScope}.
\end{itemize}

\section{Datasets}
\label{app:datasets}

This appendix details the real-world datasets used in this study: how each
was preprocessed from its source corpus into the JSON output shape we score, 
the \OA{} schemas, and
the task-native metric we report alongside the \OA{} score. The synthetic
probes (Org2Graph and \textsc{Facts2Order}) are generated rather than
preprocessed, and carry no external metric.
Because our preprocessing reshapes the original data,
the native metric we report sometimes diverges from the dataset's official
metric; we flag each such divergence in the relevant subsection.

\subsection{SciERC}
\label{app:dataset-scierc}

The raw corpus is the \texttt{sciERC\_processed} release of
\cite{scierc} (\url{http://nlp.cs.washington.edu/sciIE/}). Each raw document
carries per-sentence token lists, NER spans (token offsets), mention-pair
relations, and coreference clusters. Preprocessing turns each coreference cluster
into one entity object (Fig.~\ref{fig:scierc-sample}, middle) whose \texttt{mentions} list holds its de-duplicated
surface forms, with the cluster \texttt{type} decided by the majority vote across its
spans; NER spans in no cluster become singleton entities. Mention-pair relations
are projected to cluster-level \texttt{(subject, predicate, object)} triples and
de-duplicated, and any relation with an endpoint outside a cluster is dropped.
Token and character offsets are discarded. One training
document fails to parse, resulting in the $349/50/100$ native split. The schema scored in the reported run
(Fig.~\ref{fig:scierc-sample}, bottom) applies Hungarian alignment at two levels:
first, to match entities and then to match the mentions within each matched entity.
A coreference cluster that is wrongly split or merged, and therefore costs score.

\begin{figure*}[t]
\centering
\begin{lstlisting}[style=oacontext, aboveskip=2pt, belowskip=2pt]
Recognition of proper nouns in Japanese text has been studied as a part of the more general problem of
morphological analysis ... Our approach ... is to consider the given task as a morphological analysis problem ...
\end{lstlisting}
\oasubrule
\begin{lstlisting}[style=oajsondata, basicstyle=\ttfamily\scriptsize, aboveskip=2pt, belowskip=2pt]
{"entities": [
  {"id":"e1","type":"Task",   "mentions":[{"text":"Recognition of proper nouns"},{"text":"It"}]},
  {"id":"e3","type":"Task",   "mentions":[{"text":"morphological analysis"},{"text":"morphological analysis problem"}]},
  {"id":"e4","type":"Method", "mentions":[{"text":"`` Amorph ''"},{"text":"Amorph"},{"text":"analyzer"},{"text":"it"}]}],
 "relations": [{"subject":"e1","predicate":"PART-OF","object":"e3"}]}
\end{lstlisting}
\oasubrule
\begin{lstlisting}[style=oajsonschema, basicstyle=\ttfamily\scriptsize, aboveskip=2pt, belowskip=2pt, escapeinside={(*}{*)}]
{"type":"object", "properties": {
  "entities":  {"type":"array", "order":"align", "items": {"type":"object", "properties": {
      "id":   {"type":"string"(*\rahl{, "idScope":"entity"}*)},
      "type": {"type":"string", "score":"exact"},
      "mentions": {"type":"array", "order":"align", "items": {"type":"object", "properties": {
          "text": {"type":"string", "score":"exact"}}}}}}},
  "relations": {"type":"array", "order":"align", "items": {"type":"object", "properties": {
      "subject":   {"type":"string"(*\rahl{, "ref":"entity"}*)},
      "predicate": {"type":"string", "score":"exact"},
      "object":    {"type":"string"(*\rahl{, "ref":"entity"}*)}}}}}}
\end{lstlisting}
\caption{SciERC. \emph{Top}: the input \emph{context} (title and
abstract). \emph{Middle}: the gold output---each entity is a coreference cluster,
relations reference entities by id. \emph{Bottom}: the schema as run; the
\rahl{highlighted} fragments are exactly what referential alignment
adds---\texttt{idScope} on entity ids and \texttt{ref} on relation endpoints; the
\textbf{plain} ablation simply omits them, leaving plain string leaves compared by
literal equality.}
\label{fig:scierc-sample}
\end{figure*}

As the native metric, we use SciERC's coref-aware \emph{entity-level}
relation F1: predicted entity clusters are aligned to gold, and then relations are
scored at the cluster level and micro-averaged, with the symmetric relations
(\texttt{COMPARE}, \texttt{CONJUNCTION}) matching in either argument order.
Because preprocessing discards the token offsets and aligns clusters by
their surface mention text, this entity-level F1 is not the official span-based
SciERC relation F1, which matches entity boundaries by offset. 
The same reshaping that defines our task also redefines the metric.

\subsection{BioRED}
\label{app:dataset-biored}

The raw corpus is the BioRED release of \cite{biored}
(\url{https://ftp.ncbi.nlm.nih.gov/pub/lu/BioRED/BIORED.zip}): $600$ PubMed
titles and abstracts with document-level entity and relation annotations. 
Preprocessing closely follows SciERC's (Appendix~\ref{app:dataset-scierc}):
mentions collapse into typed, coreference-merged entities (\texttt{type} by majority
vote across the mentions; Fig.~\ref{fig:biored-sample}, middle), relations become
de-duplicated \texttt{(subject, predicate, object)} triples with any out-of-entity
endpoint dropped, offsets are discarded, and title and abstract form the input
\texttt{context} (Fig.~\ref{fig:biored-sample}, top). Two differences matter.
First, an entity is keyed by its \emph{normalized concept id}---an Entrez or MeSH
accession fixed by the source databases and not derivable from the abstract,
whereas SciERC's entity ids are arbitrary; the model must invent its own, so
relations are routed by referential alignment rather than literal-id matching
(Fig.~\ref{fig:biored-sample}, bottom), and a mention tagged with several ids
joins every matching entity. Second, the schema scores mention \texttt{text} by
\texttt{jaro\_winkler} rather than exact match and up-weights entity
\texttt{type}. We draw a seeded $50/50/100$ train/val/test subsample.

\begin{figure*}[t]
\centering
\begin{lstlisting}[style=oacontext, aboveskip=2pt, belowskip=2pt, escapeinside={(*}{*)}]
(*\textbf{Debrisoquine phenotype and the pharmacokinetics ... of metoprolol and its enantiomers.}*)
The metabolism of the cardioselective beta-blocker metoprolol is under genetic control of the
debrisoquine/sparteine type ...
\end{lstlisting}
\oasubrule
\begin{lstlisting}[style=oajsondata, basicstyle=\ttfamily\scriptsize, aboveskip=2pt, belowskip=2pt]
{"entities": [
  {"id":"D003647", "type":"ChemicalEntity",    "mentions":[{"text":"Debrisoquine"},{"text":"debrisoquine"}]},
  {"id":"154",     "type":"GeneOrGeneProduct",  "mentions":[{"text":"beta-2 receptor"},{"text":"beta-2 adrenoceptor"}]},
  {"id":"D008790", "type":"ChemicalEntity",     "mentions":[{"text":"metoprolol"}]},
  {"id":"153",     "type":"GeneOrGeneProduct",  "mentions":[{"text":"beta-1 adrenoceptor"}]}],
 "relations": [
  {"subject":"153",     "predicate":"Negative_Correlation", "object":"D008790"},
  {"subject":"D008790", "predicate":"Association",          "object":"D003647"}]}
\end{lstlisting}
\oasubrule
\begin{lstlisting}[style=oajsonschema, basicstyle=\ttfamily\scriptsize, aboveskip=2pt, belowskip=2pt, escapeinside={(*}{*)}]
{"type":"object", "properties": {
  "entities":  {"type":"array", "order":"align", "items": {"type":"object", "properties": {
      "id":   {"type":"string"(*\rahl{, "idScope":"entity"}*)},
      "type": {"type":"string", "score":"exact", "valueWeight":1.5},
      "mentions": {"type":"array", "order":"align", "items": {"type":"object", "properties": {
          "text": {"type":"string", "score":"jaro_winkler"}}}}}}},
  "relations": {"type":"array", "order":"align", "items": {"type":"object", "properties": {
      "subject":   {"type":"string"(*\rahl{, "ref":"entity"}*)},
      "predicate": {"type":"string", "score":"exact"},
      "object":    {"type":"string"(*\rahl{, "ref":"entity"}*)}}}}}}
\end{lstlisting}
\caption{BioRED. \emph{Top}: the input \emph{context}---title (bold) and
abstract. \emph{Middle}: the gold output---entities are coreference clusters
keyed by normalized concept ids (the model invents its own), relations reference
them by id. \emph{Bottom}: the schema as run; the \rahl{highlighted}
\texttt{idScope}/\texttt{ref} fragments are what referential alignment adds over
the \textbf{plain} ablation. Mention \texttt{text} uses \texttt{jaro\_winkler};
\texttt{type} is up-weighted.}
\label{fig:biored-sample}
\end{figure*}

The native metric mirrors SciERC's coreference-aware entity-level relation, F1
(Appendix~\ref{app:dataset-scierc}), but differs in two ways. BioRED relations
are directed, so there is no either-order matching (SciERC allows it for its
symmetric predicates). 
The alignment pairs entities by their surface mentions,
not their gold concept ids, so the score is not comparable to the official
id-keyed metric.

\subsection{Bio AMR}
\label{app:dataset-amrbio}

The raw corpus is Release~3.0 of the Bio AMR Corpus~\cite{bioamr}
(\url{https://amr.isi.edu/download/2018-01-25/amr-release-bio-v3.0.txt}):
${\sim}6{,}900$ sentences from cancer-related PubMed articles (each the input
\texttt{context}; Fig.~\ref{fig:amrbio-sample}, top) annotated with
Abstract Meaning Representation~\cite{amr} graphs in PENMAN notation.
Preprocessing parses each gold graph into the JSON output shape in
Fig.~\ref{fig:amrbio-sample} (middle): a \texttt{root} reference, a list of
\texttt{nodes} (each an instance carrying its \texttt{concept} symbol, with
constant-valued \texttt{attributes} nested under their node), and a flat list of
\texttt{relations} (\texttt{source}/\texttt{role}/\texttt{target}). 
The PENMAN variable letters are
dropped (they survive only as opaque node ids) so the schema routes relations
by referential alignment over those ids (Fig.~\ref{fig:amrbio-sample}, bottom). 
The conversion is lossless: every gold
round-trips back to its triple set. A single seeded shuffle with no stratification partitions the corpus
into disjoint $100/100/200$ train/val/test splits.

\begin{figure*}[t]
\centering
\begin{lstlisting}[style=oacontext, aboveskip=2pt, belowskip=2pt]
Expression and localization of PHB in pancreatic cancer cells and tissue ...
\end{lstlisting}
\oasubrule
\begin{lstlisting}[style=oajsondata, basicstyle=\ttfamily\scriptsize, aboveskip=2pt, belowskip=2pt]
{"root": "a",
 "nodes": [
  {"id":"a", "concept":"and",              "attributes":[]},
  {"id":"e", "concept":"express-03",       "attributes":[]},
  {"id":"p2","concept":"protein",          "attributes":[]},
  {"id":"n", "concept":"name",             "attributes":[{"role":":op1","value":"PHB"}]},
  {"id":"b", "concept":"be-located-at-91", "attributes":[]},
  {"id":"c", "concept":"cell",             "attributes":[]},
  {"id":"c2","concept":"cancer",           "attributes":[]}, ...],
 "relations": [
  {"source":"a", "role":":op1",  "target":"e"},
  {"source":"e", "role":":ARG2", "target":"p2"},
  {"source":"p2","role":":name", "target":"n"},
  {"source":"b", "role":":ARG1", "target":"p2"},
  {"source":"c", "role":":mod",  "target":"c2"}, ...]}
\end{lstlisting}
\oasubrule
\begin{lstlisting}[style=oajsonschema, basicstyle=\ttfamily\scriptsize, aboveskip=2pt, belowskip=2pt, escapeinside={(*}{*)}]
{"type":"object", "properties": {
  "root": {"type":"string"(*\rahl{, "ref":"node"}*)},
  "nodes": {"type":"array", "order":"align", "items": {"type":"object", "properties": {
      "id":      {"type":"string"(*\rahl{, "idScope":"node"}*)},
      "concept": {"type":"string", "score":"exact"},
      "attributes": {"type":"array", "order":"align", "items": {"type":"object", "properties": {
          "role":  {"type":"string", "score":"exact"},
          "value": {"type":"string", "score":"exact"}}}}}}},
  "relations": {"type":"array", "order":"align", "items": {"type":"object", "properties": {
      "source": {"type":"string"(*\rahl{, "ref":"node"}*)},
      "role":   {"type":"string", "score":"exact"},
      "target": {"type":"string"(*\rahl{, "ref":"node"}*)}}}}}}
\end{lstlisting}
\caption{Bio AMR. \emph{Top}: the input \emph{context}---a biomedical sentence.
\emph{Middle}: the gold output---a rooted meaning graph of \texttt{concept}
nodes, with constant attributes nested under their node and \texttt{relations}
referencing nodes by id; the variable letters are arbitrary, so the model picks
its own. \emph{Bottom}: the schema as run; the \rahl{highlighted}
\texttt{idScope}/\texttt{ref} fragments are exactly what referential alignment
adds---the \textbf{strict} ablation omits them, comparing the arbitrary variable
letters by literal equality so relation routing collapses. All other leaves are
\texttt{exact} and every array is order-agnostic (AMR is a set of triples).}
\label{fig:amrbio-sample}
\end{figure*}

As the native metric, we use AMR's standard metric, Smatch~\cite{smatch}:
the best node bijection between the predicted and gold triple sets, with F1
micro-averaged over the corpus. That bijection search is precisely \OA{}'s
referential alignment, so the registered referential schema tracks Smatch by
construction.

\subsection{NATURAL PLAN}
\label{app:dataset-natural-plan}

The raw data are the Trip Planning split of NATURAL PLAN~\cite{naturalplan}
(\url{https://github.com/google-deepmind/natural-plan}). Each
raw example ships a zero-shot prompt, the target cities and their stay durations as
\texttt{**}-delimited fields and a free-text reference plan. Preprocessing keeps the
prompt verbatim as the input \texttt{context} (Fig.~\ref{fig:natural-plan-sample}, top)
and builds the gold directly from the
\texttt{cities}/\texttt{durations} fields into
an \texttt{itinerary} list of \texttt{\{city, days\}} objects in the visiting order
(Fig.~\ref{fig:natural-plan-sample}, middle).
The NATURAL PLAN is evaluation-only (no native train split), so we keep train and val
small ($100$ each) and a $200$-example
\texttt{test} split, all difficulty-stratified by the number of cities ($3$--$10$).
The schema used in the reported run
(Fig.~\ref{fig:natural-plan-sample}, bottom) aligns the itinerary by position, so
producing the correct stays in the wrong order loses score.

\begin{figure*}[t]
\centering
\begin{lstlisting}[style=oacontext, aboveskip=2pt, belowskip=2pt]
You plan to visit 3 European cities for 14 days in total. You only take direct flights to commute between cities.
You would like to visit Florence for 6 days. You want to meet a friend in Florence between day 9 and day 14. You
would like to visit Barcelona for 5 days. You would like to visit Helsinki for 5 days.
Here are the cities that have direct flights:
Barcelona and Florence, Helsinki and Barcelona.
Find a trip plan of visiting the cities for 14 days by taking direct flights to commute between them.
\end{lstlisting}
\oasubrule
\begin{lstlisting}[style=oajsondata, basicstyle=\ttfamily\scriptsize, aboveskip=2pt, belowskip=2pt]
{"itinerary": [
  {"city":"Helsinki",  "days":5},
  {"city":"Barcelona", "days":5},
  {"city":"Florence",  "days":6}]}
\end{lstlisting}
\oasubrule
\begin{lstlisting}[style=oajsonschema, basicstyle=\ttfamily\scriptsize, aboveskip=2pt, belowskip=2pt, escapeinside={(*}{*)}]
{"type":"object", "properties": {
  "itinerary": {"type":"array", "order":(*\rahl{"fixed"}*), "items": {"type":"object", "properties": {
      "city": {"type":"string",  "score":"exact"},
      "days": {"type":"integer", "score":"invdiff"}}}}}}
\end{lstlisting}
\caption{NATURAL PLAN (Trip Planning). \emph{Top}: the input \emph{context}---the
zero-shot trip problem, stating the day budget, the per-city stays and meeting
windows, and the available direct flights. \emph{Middle}: the gold output---an
ordered \texttt{itinerary} of \texttt{\{city, days\}} stays, taken from the released
target fields. \emph{Bottom}: the schema as run; the \rahl{highlighted} value is the
\emph{sequence} arm---\texttt{order} set to \texttt{"fixed"}, which aligns the
itinerary positionally so a scrambled sequence is penalized---while the \emph{set}
ablation replaces it with \texttt{"align"} (order-blind Hungarian matching).
\texttt{city} is matched exactly (it is copied from the prompt) and \texttt{days} by
the graded inverse-difference comparator.}
\label{fig:natural-plan-sample}
\end{figure*}

As the native metric we use NATURAL PLAN's official metric, the binary
per-example exact-match solve rate: a plan solves the instance iff its
\texttt{(city, days)} itinerary reproduces the gold stay for stay, in order,
averaged over the split.

\subsection{ROCStories}
\label{app:dataset-rocstories}

The raw data are the ROCStories corpus~\cite{rocstories}, taken from 
\url{https://huggingface.co/datasets/mintujupally/ROCStories}. Each
story is split into sentences on sentence-ending punctuation, and only stories that
are segmented into exactly five sentences are retained. For each story, we draw a random
non-identity permutation and present the sentences in that scrambled order---labeled
$1..N$ in the input \texttt{context} (Fig.~\ref{fig:rocstories-sample}, top)---and recorded
as gold the permutation of those
labels that restores the original reading order (Fig.~\ref{fig:rocstories-sample}, middle). 
The splits are mutually disjoint: $100/100$ train/val and a $200$-example \texttt{test}.
The schema scored in the reported run
(Fig.~\ref{fig:rocstories-sample}, bottom) aligns the \texttt{indices} list
positionally.

\begin{figure*}[t]
\centering
\begin{lstlisting}[style=oacontext, aboveskip=2pt, belowskip=2pt]
Here are the sentences in scrambled order:

Sentence 1: Got to Austin, but had to take a plane.
Sentence 2: Jim was driving from Atlanta to Austin when his car broke down.
Sentence 3: It seemed like the middle of nowhere so he began to walk.
Sentence 4: When he returned, his car couldn't be found.
Sentence 5: He came across a gas station ten miles away and got a ride back.
\end{lstlisting}
\oasubrule
\begin{lstlisting}[style=oajsondata, basicstyle=\ttfamily\scriptsize, aboveskip=2pt, belowskip=2pt]
{"indices": [2, 3, 5, 4, 1]}
\end{lstlisting}
\oasubrule
\begin{lstlisting}[style=oajsonschema, basicstyle=\ttfamily\scriptsize, aboveskip=2pt, belowskip=2pt, escapeinside={(*}{*)}]
{"type":"object", "properties": {
  "indices": {"type":"array", "order":(*\rahl{"fixed"}*), "items": {"type":"integer", "score":"exact"}}}}
\end{lstlisting}
\caption{ROCStories (sentence ordering). \emph{Top}: the input \emph{context}---the
five story sentences presented in scrambled order and labelled $1..N$. \emph{Middle}:
the gold output---the \texttt{indices} permutation of those labels that restores the
original reading order. \emph{Bottom}: the schema as run; the \rahl{highlighted} value
is the \emph{sequence} arm---\texttt{order} set to \texttt{"fixed"}, which aligns the
\texttt{indices} list positionally so a scrambled order is penalized---while the
\emph{set} ablation replaces it with \texttt{"align"} (order-blind Hungarian matching).
The data key is the neutral \texttt{indices} (not \texttt{order}), so neither schema
nor prompt reveals that the task is a reordering.}
\label{fig:rocstories-sample}
\end{figure*}

As the native metric, we use the standard sentence-ordering
Perfect-Match Ratio (PMR)~\cite{rebart}, the
fraction of stories whose predicted order reproduces the gold permutation exactly,
alongside Kendall's $\tau$ as a graded view of partial order~\cite{rebart}. Because the predicted
bag of labels is always exactly the gold bag, the order-blind \emph{set} arm of the
\OA{} reward scores $\approx\!1.0$ regardless of sequence, so the entire
sequence-vs-set gap is attributable to order alone.

\section{Intrinsic Validation}
\label{app:intrinsic}

This appendix collects additional material for the intrinsic
validation of Section~\ref{sec:intrinsic}.

\paragraph{A worked referential-alignment instance.}
Fig.~\ref{fig:intrinsic-ra-sample} shows the unit of the invariance probe of
Section~\ref{sec:intrinsic}: a small Org2Graph instance and its relabel-only
candidate. Every identifier differs, yet referential alignment recovers the
bijection $\refmap$ and scores $1.0$, while the plain ablation compares the
renumbered identifiers by value and matches no reference. The two
\texttt{Milo}/\texttt{scientist} records show why this is nontrivial: tied in
every attribute, they are separable only by structure---twin~A is the
\emph{source} of the \texttt{reports\_to} and \texttt{knows} edges, twin~B
only a target---which the structural tie-break of Section~\ref{sec:oa-ref} uses
to pin the bijection.

\begin{figure*}[t]
\centering
\begin{minipage}[t]{0.66\textwidth}
\begin{lstlisting}[style=oajsondata]
gold:
{"people": [
   {"id":"474ee2","name":"Milo","title":"scientist"},    <- twin A
   {"id":"904d0c","name":"Milo","title":"scientist"},    <- twin B
   {"id":"478cc2","name":"Theo","title":"scientist"}],
 "companies": [
   {"id":"308da8","name":"Linton","industry":"retail"},
   {"id":"80425d","name":"Oscorp","industry":"media"}],
 "employment": [
   {"person":"474ee2","company":"308da8","role":"contractor"},
   {"person":"904d0c","company":"308da8","role":"associate"},
   {"person":"478cc2","company":"308da8","role":"associate"}],
 "acquaintance": [
   {"source":"474ee2","target":"904d0c","relation":"reports_to"},
   {"source":"474ee2","target":"478cc2","relation":"knows"}],
 "partnership": [
   {"source":"308da8","target":"80425d","relation":"competes_with"}]}
\end{lstlisting}
\end{minipage}\hfill
\begin{minipage}[t]{0.30\textwidth}
\begin{lstlisting}[style=oajsonchanges]
candidate (relabel only):
the same graph, every id
renumbered and all references
rewritten through the map:

  474ee2 -> a5cd68
  904d0c -> 4d3c1a
  478cc2 -> ca264e
  308da8 -> 18b8ff
  80425d -> 25165e
\end{lstlisting}
\end{minipage}
\caption{A small Org2Graph instance (readable values, narrow vocabulary) and its
relabel-only candidate. \emph{Every} identifier differs between gold and
candidate. The two \texttt{Milo}/\texttt{scientist} records marked twin~A
and twin~B are property-twins: identical in every attribute.}
\label{fig:intrinsic-ra-sample}
\end{figure*}

\paragraph{RA Data generation.}
Algorithms~\ref{alg:intrinsic-ra-equiv} and~\ref{alg:intrinsic-ra-perop}
provide the exact procedures for the two RA experiments described in Section~\ref{sec:intrinsic}.
In this appendix, the \textsc{emit} appends one gold/candidate pair,
with its family and magnitude tags, to the generated dataset.
Every gold draws its parameters independently via
\textsc{SampleParams} (Algorithm~\ref{alg:intrinsic-ra-equiv}, lines~2--4), so
both experiments cover the same distribution of graphs. They share three additional
primitives. \textsc{SampleGold} draws an Org2Graph instance with readable
(un-obfuscated) categorical values and the narrow $6/6/6/3/3$ vocabulary
(Section~\ref{sec:dataset-org2graph}) at the requested sizes and edge densities;
\textsc{MakeTwins} turns a fraction $t$ of each scope's records into
property-twins; \textsc{Relabel} reassigns every record a fresh identifier,
disjoint from the gold's, and rewrites all references accordingly. In
Algorithm~\ref{alg:intrinsic-ra-perop}, each $(o,k)$ candidate is built
independently from the gold. The six edit operations $\mathcal{O}$ are as follows:
\begin{itemize}
\item \emph{categorical relabel} --- pick a uniformly random categorical leaf
  (a person's title, a company's industry, an employment role, or an edge
  relation) and set it to a \emph{different} code from the same codebook
  (Section~\ref{sec:dataset-org2graph});
\item \emph{reference rerouting} --- pick a random edge endpoint whose scope
  holds ${\ge}2$ records and rewire it to a different record's id;
\item \emph{record deletion} --- pick a random scope still holding ${\ge}2$
  records, delete a random record together with every incident edge;
\item {\emph{edge deletion} --- delete a uniformly random edge;}
\item \emph{record insertion} --- insert a fresh record with a new id and
  sampled attributes (a new person also receives an employment edge to a
  random company);
\item \emph{edge insertion} --- add a new acquaintance (two distinct random
  people) or partnership (two distinct random companies) edge with a sampled
  relation.
\end{itemize}

\begin{algorithm}[t]
\caption{RA invariance (identifier relabelings): dataset generation.}
\label{alg:intrinsic-ra-equiv}
\begin{algorithmic}[1]
\REQUIRE golds $G{=}100$; relabelings per gold $K{=}5$
\ENSURE $100\times5=500$ gold/candidate pairs
\FOR{\textbf{each} of the $G$ golds}
  \STATE \textsc{SampleParams}: people $n_p \sim U\{3..10\}$, companies
         $n_c \sim U\{2..\lceil n_p/2\rceil\}$
  \STATE \quad twin density $t \sim U[0,1]$
  \STATE \quad edge densities $\rho_a \sim U[0.2,0.5]$,
         $\rho_p \sim U[0.3,0.7]$
  \STATE $g \gets \textsc{SampleGold}(n_p,n_c,\rho_a,\rho_p)$;\quad
         $\textsc{MakeTwins}(g,t)$
  \STATE emit $K$ pairs $\bigl(g,\;\textsc{Relabel}(g)\bigr)$,
         each relabeling drawn independently
\ENDFOR
\end{algorithmic}
\end{algorithm}

\begin{algorithm}[t]
\caption{RA per-operation sweep: dataset generation.}
\label{alg:intrinsic-ra-perop}
\begin{algorithmic}[1]
\REQUIRE golds $G{=}100$, parameters drawn per gold by \textsc{SampleParams}
         as in Algorithm~\ref{alg:intrinsic-ra-equiv};
         edit operations $\mathcal{O}$ ($6$ ops, see text)
\ENSURE $100\times2\times6\times9=10800$ gold/candidate pairs ($5400$ with
        the relabel, scored in Section~\ref{sec:intrinsic}; $5400$ without,
        for the control sweep)
\FOR{\textbf{each} of the $G$ golds}
  \STATE $g \gets \textsc{SampleGold}(n_p,n_c,\rho_a,\rho_p)$;\quad
         $\textsc{MakeTwins}(g,t)$
  \FOR{$r\in\{\text{relabel},\text{no-relabel}\}$,\ $o\in\mathcal{O}$,\ $k=0,\dots,8$}
    \STATE $c \gets \textsc{Relabel}(g)$ if $r{=}$ relabel, else a copy of $g$
    \FOR{$k$ times}
      \STATE \textbf{if} $o$ has no feasible target in $c$ \textbf{then break}
      \STATE apply $o$ once to $c$ at a random feasible target
    \ENDFOR
    \STATE emit $(g,\,c,\,o,\,k,\,r)$
  \ENDFOR
\ENDFOR
\end{algorithmic}
\end{algorithm}

\paragraph{RA per-operation effects.}
Table~\ref{tab:intrinsic-ra-spearman} quantifies the per-operation
sensitivity behind Fig.~\ref{fig:intrinsic-ra-perop-pooled}: the Spearman
correlation of score with the number of applications $k$ of a single edit.

\begin{table}[t]
\centering
\caption{RA per-operation sensitivity (Org2Graph generator): Spearman
correlation of score with the number of applications $k$ of a single edit,
over the sweep of Section~\ref{sec:intrinsic} ($5400$ pairs, every edit applied
on top of an identifier relabel; Algorithm~\ref{alg:intrinsic-ra-perop}).
More negative means more sensitive. In the \emph{reference rerouting} row
plain is blind to the one edit that corrupts routing alone.}
\label{tab:intrinsic-ra-spearman}
\footnotesize
\setlength{\tabcolsep}{6pt}

\begin{tabular}{l cc}
\toprule
Operation & RA & plain \\
\midrule
categorical relabel & $-0.784$ & $-0.460$ \\
reference rerouting & $\mathbf{-0.660}$ & $\mathbf{-0.011}$ \\
record deletion & $-0.818$ & $-0.776$ \\
edge deletion & $-0.771$ & $-0.627$ \\
record insertion & $-0.888$ & $-0.779$ \\
edge insertion & $-0.667$ & $-0.604$ \\
\bottomrule
\end{tabular}

\end{table}

\paragraph{A worked Facts2Order instance.}
Fig.~\ref{fig:facts2order-sample} shows a hidden-key \textsc{Facts2Order}
instance ($N{=}4$, integer key): the designated sort key (weight) is buried
among per-item numeric decoy clauses on other attributes (price, length) whose
values are unconstrained, their order shuffled per item, so the key is never
identifiable as the lone number. Only the weights are drawn distinct, fixing a
unique gold order; the optimizer must discover from \OA{} feedback that
weight---not price or length---is the sort field. (The intrinsic study described in 
Section~\ref{sec:intrinsic} ignores these surface sentences and scores the index
permutations directly; cf.\ Fig.~\ref{fig:intrinsic-order-sample}.)

\begin{figure}[t]
\centering
\footnotesize
\begin{verbatim}
Here are the items:

Item 1: It is priced at 512 dollars.
        Saiph weighs 37 kilograms.
        It measures 268 centimetres.
Item 2: Antares weighs 696 kilograms.
        It is priced at 144 dollars.
        It measures 91 centimetres.
Item 3: It measures 730 centimetres.
        Arcturus weighs 933 kilograms.
        It is priced at 305 dollars.
Item 4: Bellatrix weighs 883 kilograms.
        It measures 410 centimetres.
        It is priced at 77 dollars.

gold: {"indices": [1, 2, 4, 3]}
\end{verbatim}
\normalsize
\caption{The most complex \textsc{Facts2Order} configuration: a
hidden-key instance (the F2O-H variant of the extrinsic study,
Section~\ref{sec:results-order}). Each item states its
weight (the sort key) among numeric decoy clauses on price and length. The gold permutation
sorts the items by weight ascending.}
\label{fig:facts2order-sample}
\end{figure}

\paragraph{Order data generation.}
Algorithm~\ref{alg:intrinsic-order} provides the exact procedure for the
\textsc{Facts2Order} corruption pairs described in Section~\ref{sec:intrinsic},
which yields $4639$ pairs at the sampled sizes (the per-gold count increases with
$N$). Every gold sample
draws its parameters via \textsc{SampleParams}
(Section~\ref{sec:dataset-facts2order}), and \textsc{SampleGold} returns the unique
permutation of $1..N$ that sorts the items, which the study scores directly
(ignoring surface sentences). Each gold is corrupted once per family into a
candidate, tagged by its realized Kendall distance $d$, and all draws
use per-gold seeds for reproducibility. Each family targets a distinct failure
mode:
\begin{itemize}
\item \emph{adjacent transpositions} ($k{=}1..N{-}1$ swaps) --- walking $d$ up in approximately unit steps to trace the partial-credit curve;
  $d$ is recomputed per candidate, so a fixed $k$ spreads over a small range as shown in
  Table~\ref{tab:intrinsic-order};
\item \emph{block reversal} / \emph{block move} --- reverse, or cut and
  reinsert, one contiguous block (length ${\ge}2$): \emph{non-local} disorder
  --- large per-item displacement that a single local swap cannot produce;
\item \emph{deletion} / \emph{insertion} (drop, or inject duplicate,
  labels) --- change the candidate length, exercising the gap-aligned DP
  denominator and excess penalty. Unlike the reorderings, these alter the label
  bag, so the order-agnostic baseline reacts to them but not to pure order.
\end{itemize}
The uncorrupted gold paired with itself is the \texttt{none} ($d{=}0$)
reference row in Table~\ref{tab:intrinsic-order}. Corruptions that
reproduce the gold are resampled, because \textsc{AdjTranspose} and
\textsc{BlockMove} can compose to the identity.
Across the generated pairs, $d$ is concentrated near the gold: of the
$2864$ length-preserving pairs, it ranges $0$--$55$ with a median of $2$ (the dense
low-$d$ region comes from the adjacent-transposition dose, the sparse tail from
the block families), while the $1775$ length-changing
\emph{deletion}/\emph{insertion} pairs leave $d$ undefined.

\begin{algorithm}[t]
\caption{Order-sensitivity experiment (\textsc{Facts2Order}): dataset generation.}
\label{alg:intrinsic-order}
\begin{algorithmic}[1]
\REQUIRE golds $G{=}300$; ascending sort
\ENSURE gold/candidate pairs, each tagged by the corruption that
        produced it (\texttt{none} for the reference)
\FOR{\textbf{each} of the $G$ golds}
  \STATE draw parameters via \textsc{SampleParams}
         (Sec.~\ref{sec:dataset-facts2order})
  \STATE $g \gets \textsc{SampleGold}$
         \COMMENT{gold permutation of $1..N$}
  \STATE emit $(g,\,g)$ \COMMENT{\texttt{none}: $d{=}0$ reference}
  \FOR{$k=1,\dots,N{-}1$}
    \STATE emit $(g,\,\textsc{AdjTranspose}(g,k))$
  \ENDFOR
  \STATE emit $(g,\,\textsc{BlockReverse}(g))$
  \STATE emit $(g,\,\textsc{BlockMove}(g))$
  \FOR{$k=1,\dots,\min(3,N{-}1)$}
    \STATE emit $(g,\,\textsc{Delete}(g,k))$
  \ENDFOR
  \FOR{$k=1,\dots,3$}
    \STATE emit $(g,\,\textsc{Insert}(g,k))$
  \ENDFOR
\ENDFOR
\STATE \textbf{resample} any candidate equal to its gold
       \COMMENT{discard identity corruptions}
\end{algorithmic}
\end{algorithm}

\begin{figure}[t]
\centering
\includegraphics[width=\columnwidth]{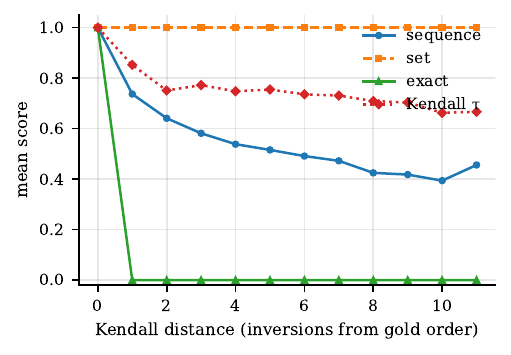}
\caption{Partial-credit curve (\textsc{Facts2Order} generator): mean score
vs.\ Kendall distance on the adjacent-transposition family. The
order-sensitive \emph{sequence} alignment decays smoothly; the order-agnostic
\emph{set} alignment is flat at $1.0$ (no order signal); \emph{exact} match is
all-or-nothing; Kendall~$\tau$ is graded here but task-specific (and $0$ on
any length change).}
\label{fig:intrinsic-order-partial}
\end{figure}

\paragraph{Order-sensitivity curves.}
Fig.~\ref{fig:intrinsic-order-partial}
shows the curve view of Table~\ref{tab:intrinsic-order}, showing the full
trend of the table samples at a few points. One feature deserves comment:
on pure reorderings Kendall~$\tau$ sits \emph{above} the sequence score at
every distance. This is a difference of scale, not of sensitivity---$\tau$
rescales the inversion \emph{count}, while the sequence alignment pays per
displaced \emph{item}---so the two are comparable in trend, but not in
level.

\clearpage

\section{System Prompts}
\label{app:prompts}

This appendix lists the initial candidate system prompt used to seed
GEPA for every dataset. As discussed in the protocol of
Section~\ref{sec:results} (paragraph ``Protocol''), each seed prompt is
deliberately minimal: beyond a one-line statement of the task and the
required JSON \emph{output} shape, it gives no guidance on \emph{how} to
produce the prediction. The task model is told what kind of object to emit, but
not the rules, edge cases, or step-by-step procedure for deriving it from the
input, so GEPA must recover that mapping purely from the \OA{}'s scalar score
and---in the feedback arm---its textual feedback. This isolates the question of whether the
metric supplies sufficient optimization ``gradient'' to drive the search from a
near-empty starting point. The prompts below are reproduced verbatim.

\begin{strip}
\paragraph{Org2Graph.}
\begin{lstlisting}[style=oaprompt]
You extract a graph of people and companies from a short English paragraph.

Output a SINGLE JSON object with exactly these five top-level keys:
  "people":       a list of {"id": <string>, "name": <string>, "title": <string>} objects.
  "companies":    a list of {"id": <string>, "name": <string>, "industry": <string>} objects.
  "employment":   a list of {"person": <person-id>, "company": <company-id>, "role": <string>} objects.
  "acquaintance": a list of {"source": <person-id>, "target": <person-id>, "relation": <string>} objects.
  "partnership":  a list of {"source": <company-id>, "target": <company-id>, "relation": <string>} objects.

Use short, stable ids of your own choosing (e.g. "p0", "p1", "c0", "c1"); the
exact id strings do not matter as long as each person / company has one
consistent id and every edge references ids defined in "people" / "companies".
"person" and acquaintance source/target are person-ids; "company" and
partnership source/target are company-ids.

Emit JSON only — no commentary, no markdown fences.
\end{lstlisting}

\paragraph{Facts2Order.}
\begin{lstlisting}[style=oaprompt]
Output a SINGLE JSON object with exactly one top-level key:
  "indices": a list of integers — a permutation of 1..N, where N is the number
             of input items. Example: {"indices": [3, 1, 4, 2]}.

Emit JSON only — no commentary, no markdown fences.
\end{lstlisting}

\paragraph{SciERC.}
\begin{lstlisting}[style=oaprompt]
You extract a scientific knowledge graph from an AI-paper abstract.

Output a SINGLE JSON object with exactly these two top-level keys:
  "entities":  a list of {"id": <string>, "type": <string>, "mentions": [{"text": <string>}, ...]} objects.
  "relations": a list of {"subject": <entity-id>, "predicate": <string>, "object": <entity-id>} objects.

Use short, stable ids of your own choosing (e.g. "e0", "e1"); the exact id
strings do not matter as long as each entity has one consistent id and every
relation's "subject" / "object" references ids defined in "entities". Each
entity carries all of its surface mentions in its "mentions" list.

Emit JSON only — no commentary, no markdown fences.
\end{lstlisting}

\paragraph{BioRED.}
\begin{lstlisting}[style=oaprompt]
You extract a biomedical relation graph from a PubMed title and abstract.

Output a SINGLE JSON object with exactly these two top-level keys:
  "entities":  a list of {"id": <string>, "type": <string>, "mentions": [{"text": <string>}, ...]} objects.
  "relations": a list of {"subject": <entity-id>, "predicate": <string>, "object": <entity-id>} objects.

Use short, stable ids of your own choosing (e.g. "e0", "e1"); the exact id
strings do not matter as long as each entity has one consistent id and every
relation's "subject" / "object" references ids defined in "entities". Each
entity carries all of its surface mentions in its "mentions" list.

Emit JSON only — no commentary, no markdown fences.
\end{lstlisting}

\paragraph{Bio AMR.}
\begin{lstlisting}[style=oaprompt]
You convert an English sentence into an Abstract Meaning Representation (AMR) graph.

Output a SINGLE JSON object with exactly these top-level keys:
  "root":      the id of the top node.
  "nodes":     a list of {"id": <string>, "concept": <string>, "attributes": [{"role": <string>, "value": <string>}, ...]} objects.
  "relations": a list of {"source": <node-id>, "role": <string>, "target": <node-id>} objects.

Use short, stable ids of your own choosing (e.g. "n0", "n1"); the exact id strings
do not matter as long as every relation and the root reference ids defined in "nodes".

Emit JSON only — no commentary, no markdown fences.
\end{lstlisting}

\paragraph{NATURAL PLAN (Trip Planning).}
\begin{lstlisting}[style=oaprompt]
You are an expert trip planner. The user message states how many days the trip
lasts, which cities to visit and for how long, any meeting/event day-window
constraints, and the available direct flights between cities. Find an itinerary
that satisfies every constraint, taking only the listed direct flights.

Output a SINGLE JSON object with exactly one top-level key:
  "itinerary": a list of {"city": <string>, "days": <integer>} objects, in the
               order the cities are visited.

Emit JSON only — no commentary, no markdown fences.
\end{lstlisting}

\paragraph{ROCStories.}
\begin{lstlisting}[style=oaprompt]
Output a SINGLE JSON object with exactly one top-level key:
  "indices": a list of integers — a permutation of 1..N, where N is the number
             of input sentences. Example: {"indices": [3, 1, 4, 2, 5]}.

Emit JSON only — no commentary, no markdown fences.
\end{lstlisting}
\end{strip}

\section{Native-Metric Results}
\label{app:native}

\begin{figure*}[t]
\centering
\includegraphics[width=\textwidth]{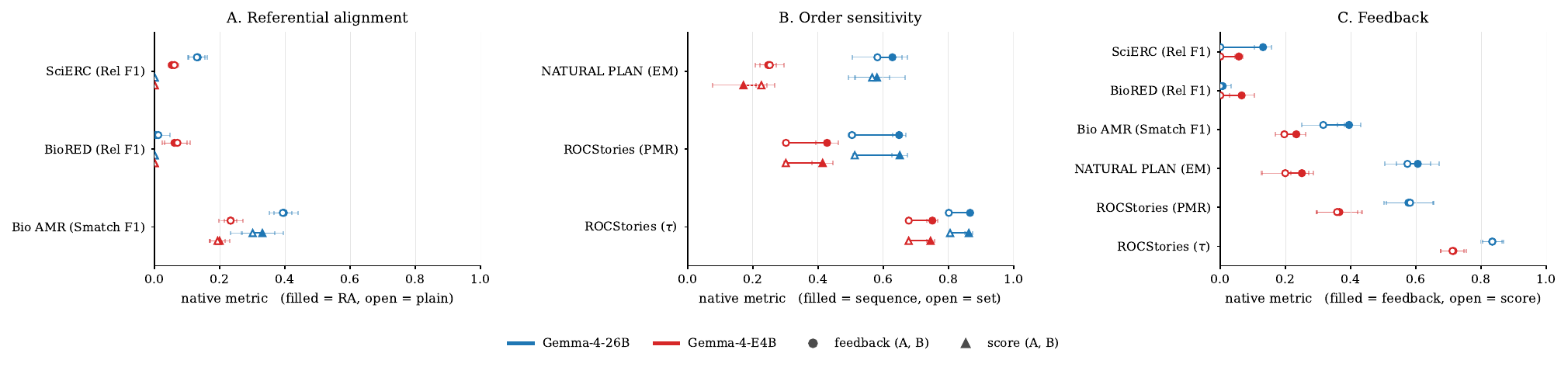}
\caption{Native-metric companion to Fig.~\ref{fig:results-forest}: the same
three contrasts and dumbbell encoding, but on each dataset's task-native
metric (annotated per row) and restricted to the real-world datasets,
since the synthetic probes carry no native metric. \textbf{(A)}~RA vs.\ plain on
relation~F1 / Smatch; \textbf{(B)}~sequence vs.\ set on EM (NATURAL PLAN)
and PMR plus Kendall's $\tau$ (ROCStories);
\textbf{(C)}~feedback vs.\ score.}
\label{fig:native-forest}
\end{figure*}

Where a real-world task carries a standard metric of its own we report it
here as a fully schema-independent cross-check on the OA-score conclusions of
Section~\ref{sec:results}. Because GEPA always optimized an \OA{} schema, a metric
that lies outside that schema (and so is unaffected by any ablation) provides an
independent external check. On all three axes, the native verdict agrees with the
headline directions of Fig.~\ref{fig:results-forest}
(Fig.~\ref{fig:native-forest}): sequence beats set
wherever order matters, feedback beats score everywhere, and on Bio~AMR,
the gain from referential alignment over the plain schema, seen in the \OA{}
score under the score reward, carries over in \emph{sign} to Smatch
($+0.030$, Gemma-4-26B), though within seed noise ($95\%$ bootstrap CI $[-0.03, +0.09]$).
One qualitative difference is
worth flagging: on SciERC and BioRED the score arm collapses to \emph{exactly}
$0.000$ relation~F1 (both RA and plain), even though its \OA{} graded score does not:
relation~F1 credits only relations whose subject, predicate, and object all match
the gold; a near-miss earns nothing. Labelling the right pair of entities
\texttt{USED-FOR}, where the gold says \texttt{PART-OF}, for instance, scores $0$
even though two of its three fields are correct. 
The model become accurate enough to register any F1 only under the feedback reward. 
Elsewhere the native and \OA{}-score gaps differ only in magnitude, agreeing in sign and
ordering.

These native metrics are not all standard leaderboard tasks. For SciERC
and BioRED, there are no published number lines up with our setup: the relation~F1 we report
is simply the conventional metric for relation extraction on data of that kind,
adopted here as a schema-independent alternative to the \OA{} score rather than
to match an external baseline. The remaining three---Bio~AMR, NATURAL PLAN, and
ROCStories---do have published reference points, which we quote below; even
there the comparison is loose (splits, models, and task framing all differ), so
we cite them only to locate the scale of each metric, not as head-to-head
results.

On Bio~AMR, the prompted model sits far below the supervised state of the
art. Our best (Gemma-4-26B, feedback RA) reaches $\approx\!0.40$ Smatch F1,
whereas purpose-built AMR parsers report low-to-mid-$0.80$s on the Bio test
set---StructBART with Maximum-Bayes-Smatch ensemble distillation~\cite{mbse} at
$\approx\!0.81$. These parsers are fine-tuned on AMR (including the Bio training
portion) and emit PENMAN on the corpus's standard split. 
We score a general-purpose open model emitting a JSON graph under a GEPA-optimized prompt
on a seeded $200$-example subsample. The gap reflects supervised, in-domain
training versus none, not a weakness of the \OA{} reward.

On NATURAL PLAN (Trip Planning), whose official metric is the exact-match
(EM) solve rate, our best Gemma-4-26B result reaches $\approx\!0.63$ EM---above the
benchmark's figures (GPT-4 $31.1\%$, the strongest model Gemini~1.5~Pro
$34.8\%$)~\cite{naturalplan}. The lead is not evidence of better planning: those
are 2024-era models (although frontier at that time) evaluated $5$-shot over the full $1600$-example set, whereas
we score a newer-generation open model on a $200$-example, city-count-stratified
split with a GEPA-optimized prompt, so the gap reflects the changed evaluation
setup (a different-generation model and an optimized prompt) not planning
ability.

On ROCStories, the prompted model again sits somewhat below the supervised state-of-the-art. 
Our Gemma-4-26B sequence arm scores $\approx\!0.65$ PMR and
$\approx\!0.87$ Kendall's $\tau$, whereas Re-BART~\cite{rebart}, fine-tuned
end-to-end to order the sentences, reports $0.82$ PMR and $0.94$ $\tau$ (full
corpus, $80{:}10{:}10$ split, $\approx\!9.8$K test stories). The gap is smaller
on the graded $\tau$ than on exact-match PMR, as expected when a prompted model
recovers the rough order but rarely the exact permutation.

\section*{Acknowledgment}
This work was supported by the Ministry of Education,
Youth and Sports of the Czech Republic through the e-INFRA CZ (ID:90254).
The author thanks Herbert Ullrich for proof-reading the manuscript and for
his valuable suggestions.

During the preparation of this work, the author used generative AI tools,
predominantly the Anthropic Claude Code (mainly Opus models)
and secondarily ChatGPT (OpenAI GPT-5). The core of the Object Aligner
library was written manually and later refactored with AI assistance.
Several extensions, the test and experimental-suite code, and supporting
utilities were drafted with AI, which was also used for code review to identify errors.
For the manuscript, AI was used for grammar correction and extensively for rephrasing.
Some passages, mainly in the related-work and dataset-description sections, were drafted by AI from detailed bullet
outlines written by the author,
and all figures and tables were produced by AI-generated scripts that construct them from the actual
computed results. AI was also used to assist in researching related topics.
The author reviewed, verified, and edited all AI-assisted content and takes full responsibility
for the content of this publication.

\bibliographystyle{IEEEtran}
\bibliography{references}

\begin{IEEEbiography}[{\includegraphics[width=1in,height=1.25in,clip,keepaspectratio]{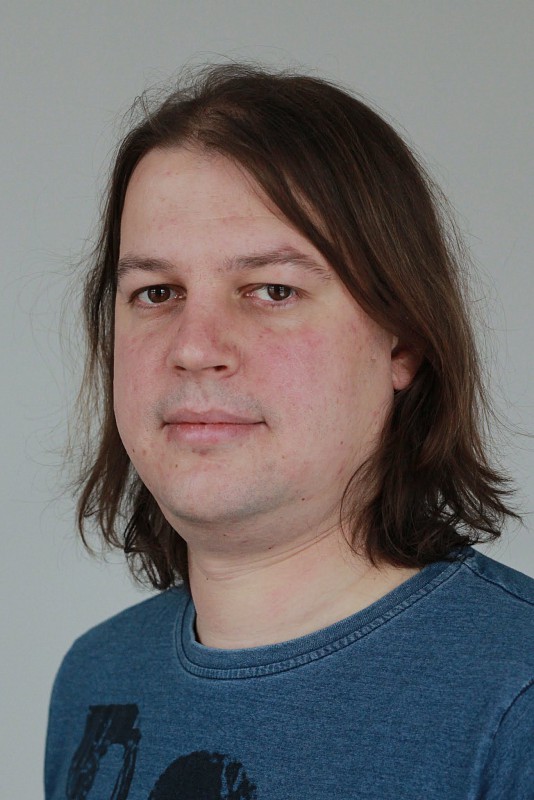}}]{Jan Drchal}
is an assistant professor at the Faculty of Electrical Engineering (FEE),
Czech Technical University in Prague (CTU). He received the Ing. (2006) and
Ph.D. (2013) degrees in electronics and computer science from CTU. His
doctoral dissertation focused on evolutionary optimization of artificial
neural networks. He is a member of the Artificial Intelligence Center
(\url{http://aic.fel.cvut.cz}).

Following his doctoral work, he worked mostly on machine learning
applications in transportation and robotics. Since 2021, his research has
centered on natural language processing, with application domains including
AI-assisted journalism, automated fact-checking, information extraction, and
prompt optimization.
\end{IEEEbiography}

\end{document}